\documentclass[journal]{IEEEtran}

\usepackage{xcolor,soul,framed} %,caption

\colorlet{shadecolor}{yellow}
\usepackage[pdftex]{graphicx}

\usepackage[cmex10]{amsmath}
\usepackage{amssymb}
\usepackage{algpseudocode}

\usepackage[ruled,vlined]{algorithm2e} 
\usepackage{soul,xcolor}
 
\usepackage{tikz}
\usetikzlibrary{positioning}
\usepackage{float}
\usepackage{booktabs}

\usepackage{array}
\usepackage{mdwmath}
\usepackage{mdwtab}
\usepackage{eqparbox}
\usepackage{url}
\usepackage{ulem}

\definecolor{darkred}{RGB}{128, 0, 34}

\newcommand{\beqa}{\begin{eqnarray}}
\newcommand{\eeqa}{\end{eqnarray}}
\newcommand{\beq}{\begin{equation}}
\newcommand{\eeq}{\end{equation}}
\newcommand{\baln}{\begin{aligned}}
\newcommand{\enaln}{\end{aligned}}
\newcommand{\bftheta}{{\mbox{\boldmath $\theta$}}}

\newcommand{\bfeta}{{\mbox{\boldmath $\eta$}}}

\newcommand{\bfSigma}{\boldsymbol\Sigma}
\newcommand{\bfPsi}{\boldsymbol\Psi}
\newcommand{\bfUpsilon}{\boldsymbol\Upsilon}

\newcommand{\0}{\textbf{0}}

\newcommand{\I}{\textbf{I}}

\newcommand{\K}{\textbf{K}}

\newcommand{\HH}{\textbf{H}}

\newcommand{\U}{\textbf{U}}

\newcommand{\Y}{\textbf{Y}}
\newcommand{\X}{\textbf{X}}

\newcommand{\T}{\mathrm{T}}

\newcommand{\uu}{\textbf{u}}

\newcommand{\ff}{\textbf{f}}
\newcommand{\vv}{\textbf{v}}

\newcommand{\xx}{\textbf{x}}
\newcommand{\yy}{\textbf{y}}

\newcommand{\ttheta}{\pmb{\theta}}

\newcommand{\bfphi}{\pmb{\phi}}

\newcommand{\bfTheta}{{\mbox{\boldmath $\Theta$}}}

\newcommand{\bflambda}{{\mbox{\boldmath $\lambda$}}}

\newcommand{\bfOmega}{{\mbox{\boldmath $\Omega$}}}
\newcommand{\bfomega}{{\mbox{\boldmath $\omega$}}}
\newcommand{\bfsigma}{{\mbox{\boldmath $\sigma$}}}
\newcommand{\calN}{\mathcal{N}}
\newcommand{\bbR}{\mathbb{R}}

\begin{document}
% \bstctlcite{IEEEexample:BSTcontrol}
    \setstcolor{red}
    \title{Sequential Estimation of Gaussian Process-based Deep State-Space Models\\
    % Deep Gaussian Process State Space models
    %\\Sequential estimation of  state space deep Gaussian processes
    %\\Sequential estimation of deep state space models with deep Gaussian processes
    %\\Sequential estimation of processes and unknown functions of deep state space models with deep Gaussian processes
    }
  \author{Yuhao Liu, %~\IEEEmembership{Student Member,~IEEE,}
      Marzieh Ajirak, %~\IEEEmembership{Student Member,~IEEE,}
      and~Petar M. Djuri\'c,~\IEEEmembership{Fellow,~IEEE}% <-this % stops a space

%   \thanks{Manuscript received July 10, 2012. \hl{This paper is an expanded paper from the IEEE MTT-S Int. Microwave Symposium held on June 17-22, 2012 in Montreal, Canada.} This work was funded in part by the Office of Naval Research under the Defense Advanced Research Projects Agency (DARPA) Microscale Power Conversion (MPC) Program under Grant N00014-11-1-0931, and in part by the Advanced Research Projects Agency-Energy (ARPA-E), U.S. Department of Energy, under Award Number DE-AR0000216.}
%   \thanks{M. Roberg is with TriQuint Semiconductor, 500 West Renner Road Richardson, TX 75080 USA (e-mail: michael.roberg@tqs.com).}% <-this % stops a space
%   \thanks{T. Reveyrand is with the XLIM Laboratory, UMR 7252, University of Limoges, 87060 Limoges, France (e-mail: tibault.reveyrand@xlim.fr).}%
%   \thanks{I. Ramos and Z. Popovic are with the Department of Electrical, Computer and Energy Engineering, University of Colorado, Boulder, CO, 80309-0425 USA (e-mail: ignacio.ramos@colorado.edu; zoya.popovic@colorado.edu).}% <-this % stops a space
%   \thanks{E. Falkenstein is with Qualcomm Inc., 6150 Lookout Road
% Boulder, CO 80301 USA (e-mail: erez.falkenstein@gmail.com).}
}  

% The paper headers
\markboth{IEEE TRANSACTIONS ON SIGNAL PROCESSING}{Liu \MakeLowercase{\textit{et al.}}: Title}

\maketitle

\begin{abstract}
%\boldmath
We consider the problem of sequential estimation of the unknowns of state-space and deep state-space models that include estimation of functions and latent processes of the models. The proposed approach relies on Gaussian and deep Gaussian processes that are implemented via random feature-based  Gaussian processes. In these models, we have two sets of unknowns, highly nonlinear unknowns (the values of the latent processes) and conditionally linear unknowns (the constant parameters of the  random feature-based  Gaussian processes). We present a method based on particle filtering where the parameters of the random feature-based Gaussian processes are integrated out in obtaining the predictive density of the states and do not need particles. We also propose an ensemble version of the method, with each member of the ensemble having its own set of features. With several experiments, we show that the method can track the latent processes up to a scale and rotation.  

\end{abstract}


\IEEEpeerreviewmaketitle

\section{Introduction}

In the last decade, the field of machine learning has seen an exceptional surge and unthinkable accomplishments \cite{dean2018new,b-goodfellow16}. {One might argue that the main reason behind its major advances has been the much improved capabilities of deep neural networks over classical machine learning methods. Enabled by their structures of  multiple processing layers, deep neural networks can learn representations of data at various levels of abstraction \cite{
%bashivan2015learning,deng2014deep, 
katircioglu2018learning,liu2017survey,   vatsa2018deep}}. 
{An area of machine learning that undergoes continued growth is Gaussian processes (GPs),} which are now routinely employed in solving hard machine learning problems. 
The reason for this is that they provide a principled, practical, and probabilistic approach to learning \cite{rasmussen2006gaussian}. Further, they are flexible, non-parametric, and computationally rather simple. They are used within a Bayesian framework that often leads to powerful methods which also offer valid estimates of uncertainties in predictions and generic model selection procedures \cite{seeger2004gaussian}. Their main drawback of computational scaling has recently been alleviated by the introduction of generic sparse approximations \cite{quinonero2005unifying,snelson2007local,titsias2009variational}.

GPs have also been used for building dynamical models \cite{beckers2021prediction}. Because of their beneficial properties, including 
bias-variance trade-off and their Bayesian
framework, they, too, have become a tool for system
identification \cite{frigola2014variational}.  The GP-based state-space models (GP-SSMs) describe dynamical systems, where one  GP models a state process \cite{kocijan2005dynamic} and another GP models the function between the states and the observations %or by a parametric structure 
\cite{beckers2021prediction}.

%\cyan{Reviewer 2 wants more discussion on particle filters and some reference to particle filters coupled with analytical filters. Reviewer 1 also added some suggested references for papers about GP state space models with kinds of filters. I agree with them. We talk very less about analytical filters. I think we can say it this way: We adopted PF here but our model can be adapted to any other analytical filters. The following contexts are the discussion on the references they provided.}

{%The use of MCMC methods is highly common for GP-SSMs. 
In the literature, there have been various approaches to inference of GP-SSMs.
%The choices of analytical filters are broadly diverse. 
For example, \cite{ko2009gp} and \cite{ko2007gp} discussed the combination of GP inference with various filters such as particle filters, extended Kalman filters, and unscented particle filters. Based on the reported results, the particle filters were generally the most accurate. However, the estimation of GPs in \cite{ko2009gp} and \cite{ko2007gp} requires the inversion of kernel matrices, which needs cubic time complexity. A computationally efficient way of GP-based inference was researched in \cite{svensson2016computationally}, and it  is based on approximating the GPs in feature spaces with numerous basis functions. The authors used particle Gibbs samplers for all the unknowns. In other words, they did not only sample the particles of the latent states but also the weight vectors and the hyperparameters, hence increasing the computational burden. Another family of efficient estimation of GP-SSMs is based on variational inference. In \cite{wang2005gaussian}, \cite{wang2007gaussian}, and \cite{eleftheriadis2017identification}, different evidence lower bounds (ELBOs) were designed and then optimized. These methods, however, are not sequential or online. In our work we adopted PF for estimating the hidden processes because this methodology is sequential in nature and has the capacity to perform estimation in highly nonlinear and nonstationary settings with any computable probability distributions.  PF has also been used in a framework where the state-transition function of a  model is parameterized using reproducing kernels \cite{tobar2015unsupervised}. Our approach in this paper  can be adapted to other types of filters.}

If the functions are described by deep mappings such as deep GPs, the resulting model is referred to as a GP-based deep state-space model (GP-DSSM) \cite{gedon2021deep}. 
{The analytical filters mentioned above are still applicable in deep state-space models (DSSMs). %Specifically, conditionally state-space models can be treated as special deep state-space models because of the hierarchical structure of the hidden processes. 
Solutions to DSSMs can be based on Rao-Blackwellized particle filters \cite{doucet2013rao,andrieu2002particle} and mixture Kalman filters  \cite{chen2000mixture}. %Although these works focus on the conditionally linear state-space models, the proposed model in this paper is open and adapt to kinds of analytical filters. 
} A subclass of DSSMs can be built by extending variational autoencoders (VAEs) as in \cite{fraccaro2018deep}.  The building blocks for these models are recurrent neural networks (RNNs) and VAEs.

Recently, methods for probabilistic  forecasting of time series based on RNNs have been proposed  \cite{rangapuram2018deep}. The objective was to learn complex patterns from raw data by an RNN combined with a parameterized  per-time-series linear state-space model. Additional efforts with similar objectives and methodologies were reported in \cite{salinas2020deepar}. In \cite{wang2019deep}, a  global-local method based on deep factor models with random effects was explored. DSSMs were also used to 
construct deep GPs  by hierarchically putting transformed GP priors on the length scales and magnitudes of the next level of GPs in the hierarchy \cite{zhao2020deep}. All these methods are different from the ones we propose here. %\cyan{There is a paper by Zhao researching the deep Gaussian process state-space model, but he used differential equations, which is different from us.}

One way of broadening the function
space of a GP is by introducing an ensemble of GPs \cite{deisenroth2015distributed,meeds2006alternative, rasmussen2002infinite,tresp2001mixtures}.   
Each GP may rely on all or on a subset of training samples and may use a unique kernel to make predictions. Ensembles of GPs have also been used for combining global approximants with local GPs \cite{snelson2007local,yuan2009variational}. In \cite{lu2020ensemble}, an ensemble of GPs was used for online interactive learning. 

We address the problem of constructing dynamic deep probabilistic latent variable models. The underlying idea is that, unlike standard state-space models, we work with DSSMs, where the variables in the intermediate layers are independently conditioned on the states from the deeper layers, and the dynamics  are generated by the process from the deepest layer, the {\em root process}. An important task of inference is the estimation of the unknowns of the model, which include the underlying parameters of the GPs and the state (latent) processes of the model.

The contributions of the paper are as follows:
\begin{itemize}
    \item a novel kernel-based method that identifies non-linear state-space systems without any information about the functions that govern the latent and observation processes,
    \item extension of the state-space models to deep structures to improve the model capacity and reveal more information about the studied phenomena, and
    \item ensemble learning to reduce the variances of the estimates of the latent processes and the predictions of the observations. 
\end{itemize}

% === II
\section{Background}
In this section, for a self-sustained presentation, we provide some background on the methodologies that are the main ingredients of the proposed solutions in this paper.

% =======
% FIG. 01
% =======
% \begin{figure}
%   \begin{center}
%   \includegraphics[width=3.5in]{pdf/01.pdf}\\
%   \caption{Microwave rectifier circuit diagram. An ideal blocking capacitor $C_b$ provides DC isolation between the microwave source and rectifying element.  An ideal choke inductor $L_c$ isolates the DC load $R_{DC}$ from RF power.}\label{circuit_diagram}
%   \end{center}
% \end{figure}

\subsection {Gaussian Processes}
A GP, written as $\mathcal{GP}\left(m(\cdot),\kappa(\cdot,\cdot|\bflambda)\right)$, is, in essence, a distribution over functions, where $m(\cdot)$ is a mean function, $\kappa(\cdot,\cdot)$ is a kernel or covariance
function, and $\bflambda$ is a vector of hyperparameters that parameterize the kernel.  
To simplify the notation, we express a GP as ${\cal GP}(m,\kappa)$ or as $\mathcal{GP}(m,\kappa(\bflambda))$, if $\bflambda$ is emphasized. For any set of inputs $\X = [\xx_j]_{j=1}^J:=[\xx_1,\ldots,\xx_J]^\top$ in the domain of a real-valued function $f \sim \mathcal{GP}(m,\kappa)$, the function values $\ff = [f(\xx_j)]_{j=1}^J$ are Gaussian distributed, i.e., 
\begin{align}
    p(\ff|\X)=\mathcal{N}(\ff|\textbf{m}_{\X},\K_{\X\X}),
\end{align}
\noindent where $\textbf{m}_{\X}=[m(\xx_j)]_{j=1}^J$ is the mean and $\K_{\X\X}:=\kappa(\X,\X|\bflambda)=[\kappa(\xx_i,\xx_j)]_{i,j}$. Given the observations $\ff$ on $\X$, the predictive distribution of  $\ff^*$ at new inputs $\X^*$ is given by \cite{rasmussen2006gaussian}
\beq
p(\ff^*|\X^*,\ff,\X)=\mathcal{N}(\ff^*|\pmb{\mu}^*,\pmb{\Sigma}^*),
\eeq
\noindent with a predictive mean and covariance obtained by 
\begin{equation}
%\label{eq:GP_prediction}
\begin{aligned}
\label{eq:10}
\pmb{\mu}^* &= \textbf{m}_{\X^*}+\K_{\X^*\X}\K_{\X\X}^{-1}(\ff-\textbf{m}_{\X}), \\
%\label{eq:11}
\pmb{\Sigma}^* &= \K_{\X^*\X^*}-\K_{\X^*\X}\K_{\X\X}^{-1}\K_{\X\X^*}.
%\notag
\end{aligned}
\end{equation}

\subsection {Random Feature-Based  Gaussian Processes}
GPs do not scale up well with $N$, the number of input-output pairs. We observe that in \eqref{eq:10}, one has to invert the $N\times N$ matrix $\K_{\X\X}$, which for large values of $N$ becomes an issue. %and \eqref{eq:11}
To ameliorate the problem, we resort to approximations by exploiting the concept of sparsity. %One approach to such an approximation is based on constructing GPs with features that come from a feature space \cite{lazaro2010sparse}.

Compared with approximations in a function space, a GP with a shift-invariant kernel has another way of approximation, one that focuses on a feature space \cite{lazaro2010sparse}. By utilizing feature spaces, the computations do not require matrix decompositions but only matrix multiplications. The vector of random features is comprised of trigonometric functions that are defined by
\begin{align}
\label{rf_vector}
\bfphi({\bf x})=\frac{1}{\sqrt{J}}&[\sin(\xx^\top\bfomega^1),\cos(\xx^\top\bfomega^1),...,\notag\\&
 \sin(\xx^\top\bfomega^J),\cos(\xx^\top\bfomega^J)]^\top,
\end{align}
where $\bfOmega=\{\bfomega^j\}_{j=1}^J$ is a set of samples randomly drawn from the power spectral density of the kernel of the GP. Then the kernel function $k(\xx,\xx')$ can be approximated by $\bfphi(\xx)^\top\bfphi(\xx')$ if the kernel is shift-invariant. It brings a type of GP approximation according to 
\beq
\label{eq:linear}
f \approx  {\bfphi(\xx)^{\top} \ttheta},
\eeq
where $\ttheta$ are parameters of the approximating model.

\subsection{Bayesian Linear Regression} 
\label{ssec:BLR}

In view of the model given by \eqref{eq:linear}, we provide a brief review of Bayesian linear regression. Consider the following model:
\beq
\label{eq:lmodel}
{y} = \bfphi^\top\ttheta + \epsilon,
\eeq
where $y$ is a scalar observation, $\epsilon$ is a zero-mean Gaussian random noise, i.e., $\epsilon\sim \calN({0}, \sigma^2)$,  with $\sigma^2$ being unknown, $\bfphi\in\mathbb{R}^{d_\theta \times 1}$ is a known feature vector, and $\bftheta \in \bbR^{d_\theta}$ is an unknown parameter vector. We assume that $\bftheta$ and $\sigma^2$ have a joint prior given by the multivariate normal--inverted Gamma distribution, i.e.,  
\begin{align}
\label{eq:NIG}
p(\bftheta, \sigma^2)\propto
\frac{1}{\sigma^{a_0+1}}
e^{-\frac{1}{2\sigma^2}\left(b_0+(\ttheta-\ttheta_0)^\top \bfSigma_0^{-1} (\ttheta-\ttheta_0)\right)},
\end{align}
where $a_0, b_0,$ $\ttheta_0$, and $\bfSigma_0$ are parameters of the prior probability density function (pdf), and where $a_0>d_\theta$ and $b_0>0$. One can show that the predictive distribution of $y$ is given by a Student's $t$-distribution \cite{b-zellner71}, that is,
\beq
\label{eq:linear_pred}
%p(y_t|\bfphi_t, \bfPhi_{t-1}, \widetilde{\yy}_{t-1}) \propto r_t^{-\frac{t-J}{2}},
p(y|\bfphi, a{_0}, b{_0}, \ttheta_0, \bfSigma{_0}) \propto \left(1+\frac{1}{\varphi_1}(y-\bfphi^\top\ttheta_0)^2\right)^{-\frac{\nu_1+1}{2}},
\eeq
where 
\begin{align}
    \label{eq:esttheta}
    \nu_1&=a_0-d_\theta,\\
    \varphi_1&=\frac{b_0}{1-\bfphi^\top \bfSigma_1 \bfphi},\\
    \label{eq:Sigma1}\bfSigma_1&=\left(\bfSigma_0^{-1}+\bfphi\bfphi^\top\right)^{-1}.
\end{align}
Thus, for the linear model in \eqref{eq:lmodel}, when the prior of  $\ttheta$ and $\sigma^2$ is given by \eqref{eq:NIG},  we have an analytical expression for the predictive distribution of $y$.

For the posterior of $\ttheta$ and $\sigma^2$
we have
\begin{align}
&p(\ttheta,\sigma^2|y,\bfphi, a_0, b_0, \ttheta_0, \bfSigma_0)\notag\\
&\propto
\frac{1}{\sigma^{a_1+1}}e^{-\frac{1}{2\sigma^2}\left(b_1+(\ttheta-\widehat{\ttheta}_1)^\top \bfSigma_1^{-1} (\ttheta-\widehat{\ttheta}_1)\right)},
\end{align}
where
\begin{align}
\label{eq:aparam}
a_1&=a_0+1,\\
\label{eq:bparam}
b_1&=b_0+y^2+\bftheta_0^\top\bfSigma_0^{-1}\bftheta_0 -
\widehat{\bftheta}_1^\top\bfSigma_1^{-1}\widehat{\bftheta}_1,\\
\label{eq:theta1}
\widehat{\ttheta}_1&= \bfSigma_1\left(\bfSigma_0^{-1}\ttheta_0 + \bfphi y\right).
\end{align}
Clearly, the posterior pdf is also a multivariate normal--inverse Gamma pdf with parameters $a_1, b_1, \ttheta_1$, and $\bfSigma_1$, which are updated from $a_0, b_0, \ttheta_0$ and $\bfSigma_0$ using \eqref{eq:aparam}, \eqref{eq:bparam}, \eqref{eq:theta1} and \eqref{eq:Sigma1}, respectively.

\subsection {Particle Filtering}
In the proposed approach, we will use concepts from particle filtering theory, and in this subsection, we provide the basics of it. {Particle filters have the capacity to work sequentially with highly nonlinear models.} In many signal processing problems, we aim at tracking a latent process $\xx_t \in \mathbb{R}^{d_x}$ of a state-space model given by
\begin{align}
    {\rm transition\;pdf:} &\;\;\; p(\xx_t|\xx_{t-1}),\\
    {\rm likelihood\;of\;}\xx_t: &\;\;\;p(y_t|\xx_t),
\end{align}
where $t$ is a discrete-time index, and $y_t\in\mathbb{R}$ is an observation process. Typically, the main objective of PF is to obtain the filtering pdf $p(\xx_t|y_{1:t})$ from $p(\xx_{t-1}|y_{1:t-1}).$

In brief, particle filters approximate the pdfs of interest by discrete random measures, where the support of a pdf is given by a set of particles and where each particle is given a weight following fundamental principles. PF is implemented as follows \cite{arulampalam2002tutorial,djuric2003particle, doucet2009tutorial}. Suppose that at time $t-1$ the filtering density $p(\xx_{t-1}|y_{1:t-1})$ is approximated by
\begin{align}
\label{eq:approx}
    {p}^M(\xx_{t-1}|y_{1:t-1})&=
    \frac{1}{M}\sum_{m=1}^M \delta(\xx_{t-1}-\xx_{t-1}^{(m)}),
\end{align}
where the symbol $\xx_{t-1}^{(m)}$ represents the $m$th particle (sample) of $\xx_{t-1}$, $\delta(\cdot)$ is the Dirac delta function, and $M$ is the number of particles. Then we can obtain ${p}^M(\xx_{t}|y_{1:t})$ from ${p}^M(\xx_{t-1}|y_{1:t-1})$
by implementing three steps:
\begin{enumerate}
    \item Generate particles $\xx_t^{(m)}$ from the predictive pdf of $\xx_t$, i.e.,
    \begin{align}
        \xx_t^{(m)}&\sim p(\xx_t|\xx_{t-1}^{(m)}).
    \end{align}
    \item Compute the weights of the particles $\xx_t^{(m)}$ according to the likelihood of $\xx_t$, or
    \begin{align}
    w_t^{(m)}&\propto p(y_t|\xx_t^{(m)}),
    \end{align}
    and where 
    \begin{align}
    \sum_{m=1}^M w_t^{(m)}&=1.
    \end{align}
    The approximation of $p(\xx_{t}|y_{1:t})$ is then given by
    \begin{align}
    {p}^M(\xx_{t}|y_{1:t})= \sum_{m=1}^M w_t^{(m)}\delta(\xx_t-\xx_t^{(m)}).
    \end{align}
    \item Resample the particles using their weights $w_t^{(m)}$ and construct a posterior of $\xx_{t}$ with equal weights and where some of the particles are replicated \cite{li2015resampling}. 
\end{enumerate}

\section{Gaussian Process State Space Model}
\begin{figure}
\small
    \centering
\begin{tikzpicture}
[
roundnode/.style={circle, draw=black, minimum size=0.7 cm},
dot/.style={circle, draw=white,  minimum size=0.7cm},
]
%Nodes
\node[roundnode]   (p1)    {$\xx_{0}$};
% \node[roundnode,fill=lightgray]     (p2)     [below=0.5cm of p1] {$\yy_{0}$};

\node[roundnode]   (s1)    [right=0.5cm of p1]{$\xx_{1}$};
\node[roundnode,fill=lightgray]   (s2)     [below=0.5cm of s1] {$\yy_{1}$};

\node[dot]     (q1)       [right=0.5cm of s1] {$\dots$};
\node[dot]     (q2)    [below=0.5cm of q1] {$\dots$};

\node[roundnode]     (r1)    [right=0.5cm of q1] {$\xx_{T}$};
\node[roundnode,fill=lightgray]     (r2)    [below=0.5cm of r1] {$\yy_{T}$};

%Lines
% \draw[->] (p1.south) -- (p2.north);
\draw[-To] (p1.east) -- (s1.west);
\draw[dotted,-To] (q1.south) -- (q2.north);
\draw[->] (s1.south) -- (s2.north);

\draw[->] (s1.east) -- (q1.west);
\draw[->] (r1.south) -- (r2.north);
\draw[->]  (q1.east) -- (r1.west);
% \draw[->] (r1.east) -- (4.5, 0);
% \draw[->] (-1, 0) -- (point.west);
\end{tikzpicture}
\caption{A generic diagram of an SSM.}
\end{figure}
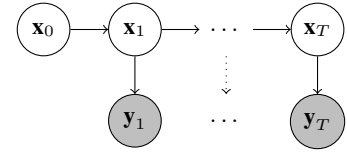

Now we introduce the GP-based state space model. Suppose the observation process $\yy_t \in \mathbb{R}^{d_y}$ is produced by a state-space model defined by
\begin{align}
%{\rm transition\;pdf:} &\;\;\; 
\label{eq:state}
\xx_t &= f(\xx_{t-1}) + {\bf u}_t,\\
%{\rm likelihood\;of\;}f_t: &\;\;\;
\label{eq:obs}
\yy_t &= g(\xx_t) + {\bf v}_t,
\end{align}
where \eqref{eq:state} represents the latent state transition equation with the state vector $\xx_t \in \mathbb{R}^{d_x}$ at time instant $t$, and \eqref{eq:obs} is the observation equation  with $\yy_t\in\mathbb{R}^{d_y}$ being the vector of observations at time instant $t$. The symbols $\uu_t \sim \mathcal{N}(\0, \sigma_u^2\I)$ and $\vv_t \sim \mathcal{N}(\0, \sigma_v^2\I)$ represent Gaussian distributed errors (noises). A generic graphical representation of an SSM is shown in Fig. 1.

Next, we express the above two equations using random feature-based GPs. In that case, we write them according to
\begin{align}
%{\rm transition\;pdf:} &\;\;\; 
\label{eq:ssm_trans}
\xx_t &= \HH^\top\bfphi^x(\xx_{t-1}) + \uu_t,\\
%{\rm likelihood\;of\;}f_t: &\;\;\;
\label{eq:ssm_obs}
\yy_t &= \bfTheta^\top\bfphi^y(\xx_t) + \vv_t,
\end{align}
where the parameters are given by the elements of the matrices $\HH\in\mathbb{R}^{2J_x\times d_x}$, $\HH = [\bfeta^{[1]},\bfeta^{[2]},\ldots,\bfeta^{[d_x]}]$, and $\bfTheta\in\mathbb{R}^{2J_y\times d_y}$, $\bfTheta = [\bftheta^{[1]},\bftheta^{[2]},\ldots,\bftheta^{[d_y]}]$. The parameters  $\bfeta^{[i]}$ and $\bftheta^{[j]}$ are initialized by Gaussian priors and updated by following  Bayesian rules.
Thus, each dimension of ${\bf x}_t$ and ${\bf y}_t$ is modeled by its own set of parameters. Further, note that the feature vectors  $\bfphi^x\in \mathbb{R}^{2J_x}$ and $\bfphi^y\in\mathbb{R}^{2J_y}$ in \eqref{eq:ssm_trans} and \eqref{eq:ssm_obs} are different because they are defined by different sets of samples, 
%embedded random features 
$\bfOmega^x$ and $\bfOmega^y$, respectively. To  simplify the notation, we use  $\bfphi^x(\xx_{t-1})=:\bfphi_{t-1}^x$ and $\bfphi^y(\xx_{t})=:\bfphi_{t}^y$. We assume that the parameter variables are all independent, i.e., the columns of $\HH$ and $\bfTheta$ are independent of the remaining columns. The noises ${\bf u}_t$ and ${\bf v}_t$ {are i.i.d. zero-mean Gaussians}, where ${\bf u}_t\sim {\cal N}({\bf 0}, {\bfSigma}_u)$ and ${\bf v}_t\sim {\cal N}({\bf 0}, {\bfSigma}_v)$, with ${\bfSigma}_u={\rm diag}(\sigma_{u}^{2{[1]}}, \sigma_{u}^{2{[2]}},\ldots, \sigma_{u}^{2{[d_x]}})$ and 
${\bfSigma}_v={\rm diag}(\sigma_{v}^{2{[1]}}, \sigma_{v}^{2{[2]}},\ldots, \sigma_{u}^{2{[d_y]}})$. 
%The independence assumption about the parameter variables implies that the dimensions of $\xx_t$ and $\yy_t$ are independent. 

The model described by \eqref{eq:ssm_trans} and \eqref{eq:ssm_obs} contains many unknowns, that is, the vector processes $\xx_t$, $t=1, 2, \ldots$, the parameter matrices $\HH$ and $\bfTheta$, and the noise variances $\sigma_u^{2[d]}$, $d=1, 2, \ldots. d_x$ and $\sigma_v^{2[d]}$, $d=1, 2, \ldots. d_y$. Conditioned on $\xx_t$, the model is of the same form as the one in \eqref{eq:lmodel}, whereas conditioned on $\HH$ and $\bfTheta$, the model given by \eqref{eq:ssm_trans} and \eqref{eq:ssm_obs} is very nonlinear in $\xx_t$.  
On account of the intractable analytical inference, we resort to PF to estimate sequentially %the parameters and 
the latent states. Given the estimated states, we update %To do the sequential inference on 
the  joint distributions of
$\HH$ and ${\bfSigma}_u$ and of  $\bfTheta$ and ${\bfSigma}_v$, respectively. %and $p(\xx_0)$ 
For these updates, we apply Bayesian linear regressions, 
%by applying \eqref{eq:linear_pred} to \eqref{eq:ssm_trans} and \eqref{eq:ssm_obs}
where we use multivariate normal--inverse Gamma pdfs for the joint priors of  $(\bfeta^{[d]}, \sigma_u^{2[d]})$ and $(\bftheta^{[d]}, \sigma_v^{2[d]})$, respectively. 

Next, we explain how we implement the following:
\begin{enumerate}
\item the propagation of  ${\bf x}_t$,
    \item the updating of the joint posteriors of $\bfeta^{(m),[d]}$ and $\sigma_u^{{(m),[d]}^2}$, for $d=1, 2, \ldots, d_x$, $m=1, 2, \ldots, M$,
    \item the updating of the joint posteriors of $\ttheta^{(m),[d]}$ and $\sigma_v^{{(m),[d]}^2}$, for $d=1, 2, \ldots, d_y$, $m=1, 2, \ldots, M$, and
    \item the weight computation of the particles and the estimation of ${\bf x}_t$.    
\end{enumerate}  
Suppose that before propagating the samples of the latent process at time $t-1$, we have $M$ particles of ${\bf x}_{t-1}$, ${\bf x}_{t-1}^{(m)}$, $m=1, 2, \ldots, M$. Assume also that for each stream of particles $m$ at $t-1$ we have the joint posterior of $\bfeta^{[d]}$ and $\sigma_u^{2[d]}$, which is a multivariate normal--inverted Gamma pdf with parameters $a_{t-1}^x, b^{(m),[d]}_{t-1}, {\bfeta}^{(m),[d]}_{t-1}$  and $\bfPsi^{(m),[d]}_{t-1}$. Further, we have the joint posterior of $\ttheta^{[d]}$ and $\sigma_v^{2[d]}$, which is also a multivariate normal--inverted Gamma pdf and with parameters $a_{t-1}^y, c^{(m),[d]}_{t-1}, {\ttheta}^{(m),[d]}_{t-1}$ and $\bfUpsilon^{(m),[d]}_{t-1}$.

\subsection{Propagation of the particles}
 We generate the elements of the particles $\xx_t^{(m)}$, $\xx_t^{(m),[d]}$, from respective univariate Student's $t$-distributions given by (see also \eqref{eq:linear_pred})
\begin{align}
&p(x_t^{(m),[d]}|\xx_{t-1}^{(m)}, \Y_{t-1}) \notag\\&
\propto \left( 1+\frac{1}{\psi_t^{(m),[d]}}\left(x_t^{(m),[d]} - \alpha_t^{(m),[d]}\right)^2 \right)^{-\frac{\nu_{t-1}^x+1}{2}},
\label{eq:transit_unknowns-1n}
\end{align}
where $d=1, 2, \ldots, d_x$, $m=1, 2, \ldots, M,$ and
\begin{align}
\label{eq:parameters_unknown-nu}
\nu_{t-1}^x& =a_{t-1}^x-2J_x,\\
\alpha_t^{(m),[d]} &= \bfphi_{t-1}^{{x^{(m)}}^\top}{\bfeta}_{t-1}^{(m),[d]},\\
\label{eq:psi}
\psi_t^{(m),[d]} &= \frac{b_{t-1}^{(m),[d]}}{1-\bfphi_{t-1}^{{x^{(m)}}^\top}\bfPsi_t^{(m),[d]}\bfphi_{t-1}^{{x^{(m)}}}},\\
\label{eq:upPsi}
\bfPsi_t^{(m),[d]}&=\left(\bfPsi_{t-1}^{{(m),[d]}^{-1}}+\bfphi_{t}^{{x^{(m)}}}\bfphi_{t}^{{x^{(m)}}^\top}\right)^{-1}.
\end{align}
Thus, the propagation  includes generating  particles ${\bf x}_t$ by \eqref{eq:transit_unknowns-1n}. For each dimension of ${\bf x}_t$, we sample $M$ particles (thus, we have a total of $Md_x$ particles), and they represent the support of ${\bf x}_t$.

\subsection{Updating of the joint posteriors of $(\bfeta^{(m),[d]}, \sigma_u^{{(m),[d]}^2})$}

The joint posterior of $(\bfeta^{(m),[d]}, \sigma_u^{{(m),[d]}^2})$ is a multivariate normal--inverted Gamma pdf with parameters $a_t^x, b_t^{(m),[d]}, \bfeta_t^{(m),[d]},$ and $ \bfPsi_t^{(m),[d]}.$ We update $\bfPsi_{t-1}^{(m),[d]}$ by \eqref{eq:upPsi}, and we find the remaining parameters recursively by  
\begin{align}
\label{eq:aparamx}
a_t^x&=a_{t-1}^x+1,\\
\label{eq:bparamx}
b_t^{(m),[d]}&=b_{t-1}^{(m),[d]}+\left(x_t^{(m),[d]}\right)^2\notag\\
&+\bfeta_{t-1}^{{(m),[d]}^\top}\bfPsi_{t-1}^{{(m),[d]}^{-1}}\bfeta_{t-1}^{(m),[d]}\notag\\
&-\bfeta_{t}^{{(m),[d]}^\top}\bfPsi_{t}^{{(m),[d]}^{-1}}\bfeta_{t}^{{(m),[d]}^\top},\\
\label{eq:eta1}
{\bfeta}_t^{(m),[d]}&= \bfPsi_t^{(m),[d]}\left(\bfPsi_{t-1}^{{(m),[d]}^{-1}}\bfeta_{t-1}^{(m),[d]} + \bfphi_{t}^{{x^{(m)}}} x_t^{(m),[d]}\right).
\end{align}

\subsection{Updating of the joint posteriors of $(\ttheta^{(m),[d]}, \sigma_v^{{(m),[d]}^2})$}

The proposed method also requires updating of the joint posteriors of $\ttheta^{(m),[d]}$ and $\sigma_v^{{(m),[d]}^2}$ for $m=1, 2, \ldots, M$, and $d= 1, 2, \ldots, d_y$.    
The joint posterior of $(\bfeta^{(m),[d]}, \sigma_u^{{(m),[d]}^2})$ is a multivariate normal--inverted Gamma pdf with parameters $a_t^y, c_t^{(m),[d]}, \bfeta_t^{(m),[d]},$ and $ \bfUpsilon_t^{(m),[d]}.$ Upon receiving ${\bf y}_t$, these parameters are updated by  
\begin{align}
\label{eq:aparam_2}
a_t^y&=a_{t-1}^y+1,\\
\label{eq:bparam_2}
c_t^{(m),[d]}&=c_{t-1}^{(m),[d]}+\left(y_t^{[d]}\right)^2\notag\\
&+\ttheta_{t-1}^{{(m),[d]}^\top}\bfUpsilon_{t-1}^{{(m),[d]}^{-1}}\ttheta_{t-1}^{(m),[d]}\notag\\
&-\ttheta_{t}^{{(m),[d]}^\top}\bfUpsilon_{t}^{{(m),[d]}^{-1}}\ttheta_{t}^{{(m),[d]}^\top},\\
\label{eq:theta1_2}
{\ttheta}_t^{(m),[d]}&= \bfUpsilon_t^{(m),[d]}\left(\bfUpsilon_{t-1}^{{(m),[d]}^{-1}}\ttheta_{t-1}^{(m),[d]} + \bfphi_{t}^{{y^{(m)}}} y_t^{[d]}\right),\\
\label{eq:upUpsilon_2}
\bfUpsilon_t^{(m),[d]}&=\left(\bfUpsilon_{t-1}^{{(m),[d]}^{-1}}+\bfphi_{t}^{{y^{(m)}}}\bfphi_{t}^{{y^{(m)}}^\top}\right)^{-1}.
\end{align}

\subsection{Weight computation of particles and estimation of ${\bf x}_t$}

We need to assign  weights to each particle $\xx_t^{(m)}$ according to  the likelihood of ${\bf x}_t^{(m)}$. The computation proceeds according to
\beq
\label{eq:weight_pf}
\widetilde{w}_t^{(m)} =  p(\yy_t|\xx_t^{(m)}, \X_{t-1}^{(m)},\Y_{t-1}),
\eeq
where $p(\yy_t|\xx_t^{(m)}, \X_{t-1}^{(m)}, \Y_{t-1})$ is the likelihood of $\xx_t^{(m)}$ given $\yy_t$, $\X_{t-1}^{(m)}$, and $\Y_{t-1}$, and where $\X_{t-1}^{(m)}$ represents all the particles generated in the $m$th stream up to time instant $t-1$, $\Y_{t-1}$ stands for all the vector observations up to time instant $t-1$, and $\widetilde{w}_t^{(m)}$ is the non-normalized weight of ${\bf x}_t^{(m)}$.

We obtain the likelihood by exploiting \eqref{eq:ssm_obs}, where we use the made assumption that ${\bf v}_t$ is Gaussian. We find that $p(\yy_t|\xx_t^{(m)}, \X_{t-1}^{(m)}, \Y_{t-1})$ is a product of $d_y$ Student's $t$-distributions, i.e.,    
\begin{align}
    &p(\yy_t|\xx_t^{(m)}, \X_{t-1}^{(m)},\Y_{t-1})\notag\\
    &\propto
    \label{eq:parameters35}
    \prod_{d=1}^{d_y}\left( 1+\frac{1}{\upsilon_t^{(m),[d]}}\left(y_t^{[d]} - \beta_t^{(m),[d]}\right)^2 \right)^{-\frac{\nu_{t-1}^y+1}{2}},
\end{align}
where $d=1, 2, \ldots, d_y$, $m=1, 2, \ldots, M,$ and
\begin{align}
\label{eq:parameters_nu}
\nu_{t-1}^y& =a_{t-1}^y-2J_y,\\
\beta_t^{(m),[d]} &= \bfphi_{t}^{{y^{(m)}}^\top}{\ttheta}_{t-1}^{(m),[d]},\\
\label{eq:upsi}
\upsilon_t^{(m),[d]} &= \frac{c_{t-1}^{(m),[d]}}{1-\bfphi_{t}^{{y^{(m)}}^\top}\bfUpsilon_{t}^{(m),[d]}\bfphi_{t}^{{y^{(m)}}}}. 
\end{align}
Once we compute the non-normalized weights by \eqref{eq:weight_pf}, we normalize them according to 
\begin{align}
\label{eq:weightnorm}
w_t^{(m)}&=\frac{\widetilde{w}_t^{(m)}}{\sum_{k=1}^M \widetilde{w}_t^{(k)}}.
\end{align}
After normalizing the weights, the minimum mean square estimate (MMSE) of $\xx_t$ is obtained by
\beq
\label{eq:mmse_pf-1}
\widehat{\xx}_t = \sum_{m=1}^M w_t^{(m)}\xx_t^{(m)}.
\eeq
The approximation of the posterior $p(\xx_t|\Y_t)$ is then given by
\beq
\label{eq:posterior_pf}
p^M\left(\xx_t|\Y_t\right) = \sum_{m=1}^M w_t^{(m)}\delta\left(\xx_t-\xx_t^{(m)}\right).
\eeq
Finally, we resample $M$ particles $\xx_t^{(m)}$ from $p^M(\xx_t|\Y_t)$ to obtain the particles that will be used for propagation in the next time instant $t+1$.

The complete procedure is summarized by Algorithm \ref{alg:pf}. We point out that an alternative algorithm can be applied where all the particles share the same parameters $\HH$ and $\bf\Theta$.

\begin{algorithm}[htbp]
\caption{Single Sequential GP-SSM}
\label{alg:pf}
\For{$m=1$ to $M$}{
    Sample $\xx_{1}^{(m)} \sim p(\xx_{1})$;\\
    Initialize the weights of $\xx_{1}^{(m)}$ as $w_1^{(m)}=1/M$, $\forall m$;
}
\For{$t=2$ to $T$ }{
    \textbf{Propagation of the states:} \\
    Sample $\xx_{t}^{(m)}$ according to \eqref{eq:transit_unknowns-1n};\\
    \textbf{Updating the parameters of the joint posterior of $(\bfeta^{(m),[d]}, \sigma_u^{{(m),[d]}^2})$:} \\
    Update $a_t^x, b_t^{(m),[d]}, \bfeta_t^{(m),[d]},$ and $ \bfPsi_t^{(m),[d]}$ via \eqref{eq:aparamx},   \eqref{eq:bparamx}, \eqref{eq:eta1}, and \eqref{eq:upPsi}, $\forall d$ and $m$;\\
    \textbf{Updating the parameters of the joint posterior of $(\ttheta^{(m),[d]}, \sigma_v^{{(m),[d]}^2})$:} \\
    Update $a_t^y, c_t^{(m),[d]}, \ttheta_t^{(m),[d]},$ and $ \bfUpsilon_t^{(m),[d]}$ via \eqref{eq:aparam_2},   \eqref{eq:bparam_2}, \eqref{eq:theta1_2}, and \eqref{eq:upUpsilon_2}, $\forall d$ and $m$;\\
\textbf{Weight computation and normalization:} \\
    Compute the weights of $\xx_t^{(m)}$ according to \eqref{eq:weight_pf} and normalize them by \eqref{eq:weightnorm}.\\
    \textbf{Estimation of the state:} \\
    Estimate $\xx_t$ by \eqref{eq:mmse_pf-1}.\\
\textbf{Resampling:} \\
    Resample $\xx_t^{(m)}$ based on their weights.
}
\end{algorithm}

% === IV
\section{Ensemble Learning}
The use of only a single set of random samples, $\bfOmega$, might not be sufficiently accurate.  In order to mitigate the problem, we introduce an ensemble of different sets of $\bfOmega$ and then combine the results obtained by each set. 
{Let $\kappa^s$ be a shift-invariant kernel from a known kernel dictionary $K:= \{\kappa^1,..., \kappa^S\}$. Ideally, $K$ should be built as large as computational resources allow. We create the sets $\bfOmega^s$ by sampling from the power spectral density of each kernel candidate $\kappa^s$. 
For estimating the latent state, we use these sets as follows.
If $\bfOmega^s$ is the $s$th set, the posterior contribution or weight of the GP based on the $s$th set to the estimate of the latent state at time $t$ is $w^s_t\propto p(s|\Y_t)$. Then, the predictive density of $\yy_t$ at time $t$
is obtained from
\beq
\begin{aligned}
p(\yy_t|\Y_{t-1})
&= \sum_{s=1}^{S}p(s|\Y_{t-1}) p(\yy_t|s,\Y_{t-1}) \\
&= \sum_{s=1}^{S}w^s_{t-1} p(\yy_t|s,\Y_{t-1}),
\end{aligned}
\eeq
where $S$ is the total number of sets and where the posterior weight is updated by
\beq
\begin{aligned}
w_t^s
= \frac{p(s|\Y_{t-1}) p(\yy_t|s,\Y_{t-1})}{p(\yy_t|\Y_{t-1})}
\propto w_{t-1}^sp(\yy_t|s,\Y_{t-1}).
\end{aligned}
\eeq

\subsection{Ensemble Estimates of the States}
\label{sec:ens_states}
The ensemble estimate of the latent states is given by the mixture
\beq
\begin{aligned}
p(\widehat{\xx}_t|\Y_t)
% &= \sum_{s=1}^{S}p(s|\Y_t) p(\widehat{\xx}_t|s,\Y_t) \\
&= \sum_{s=1}^{S}w^s_{t} p(\widehat{\xx}_t|s,\Y_t).
\end{aligned}
\eeq
We point out that the estimates of the latent states by random feature-based methods are identifiable up to a scale, shift, and rotation \cite{gundersen2021latent}. Thus, to fuse  the state estimates, we have to force the estimators of all the ensemble members into the same coordinate base. To that end, we arbitrarily fix the rotation of $\X\in \mathbb{R}^{T\times d_x}$ by taking the singular value decomposition (SVD) of the MMSE estimate, $\widehat{\X} = \U\textbf{S}\textbf{V}^\top$, and setting the new estimated $\widehat{\X}\in \mathbb{R}^{T\times d_x}$ as the columns of the left singular vectors $\textbf{U}$ with $d_x$ largest singular values. Then  we mirror and rotate all the candidate latent states so that they have the same pattern. First, we set a guidance point $\Tilde{\xx}_{t}$ with respect to a specific time $t$. Then we rotate all the latent states $\Tilde{\X}^{(s)}$ to make sure that $\Tilde{\xx}_{t}^{(s)}$ is ``overlapped'' with $\Tilde{\xx}_{t}$. Finally, we take the weighted average of $\Tilde{\X}^{(s)}$ as the ensemble estimate of  $\Tilde{\X}$. 

\subsection{Keep and Drop}
We use the individual estimates of the ensemble members to improve on their respective estimates. 
In practice, if we do not take precautionary measures, only a small portion of them would remain with significant weights. 
For this reason, we remove the members with small weights using the principle of resampling  
and replace them with members that perform much better. With replacements, we reduce the diversity of features in the ensemble but increase the number of particles that explore the spaces of the latent processes with the features of the replicated members. 
Further, we note that candidate models need to be trained at the beginning. For this stage, we fix the weights to $w_t^s \equiv 1/S$ at the beginning until $t=\T_0$. 

\section{Gaussian Process-based Deep State-Space Models}
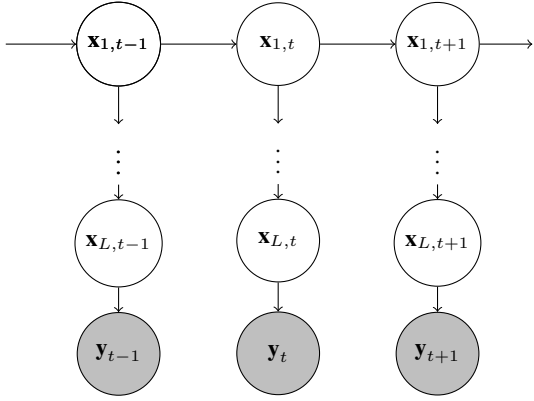
\begin{figure}
\small
    \centering
\begin{tikzpicture}
[
roundnode/.style={circle, draw=black, minimum size=1.1 cm},
squarednode/.style={rectangle, draw=black,  minimum size=5mm},
dot/.style={circle, draw=white,  minimum size=0.7cm},
]
%Nodes
\node[roundnode]   (point)    {$\xx_{1, t-1}$};
\node[dot]         (p2)        [below=0.5cm of point] {$\vdots$};

\node[roundnode]   (point)    {$\xx_{1, t-1}$};
\node[dot]         (p2)        [below=0.5cm of point] {$\vdots$};
\node[roundnode]     (p3)     [below=1.5cm of point] {$\xx_{L, t-1}$};
\node[roundnode,fill=lightgray]     (p4)     [below=3cm of point] {$\yy_{t-1}$};

\node[roundnode]     (q1)       [right=of point] {$\xx_{1, t}$};
\node[dot]            (q2)      [below=0.5cm of q1] {$\vdots$};
\node[roundnode]      (q3)     [below=1.5cm of q1] {$\xx_{L, t}$};
\node[roundnode,fill=lightgray]      (q4)    [below=3cm of q1] {$\yy_{t}$};

\node[roundnode]     (r1)    [right=of q1] {$\xx_{1, t+1}$};
\node[dot]           (r2)    [below=0.5cm of r1] {$\vdots$};
\node[roundnode]     (r3)    [below=1.5cm of r1] {$\xx_{L, t+1}$};
\node[roundnode,fill=lightgray]     (r4)    [below=3cm of r1] {$\yy_{t+1}$};

%Lines
\draw[->] (point.south) -- (p2.north);
\draw[->] (p2.south) -- (p3.north);
\draw[->] (p3.south) -- (p4.north);

\draw[->] (point.east) -- (q1.west);

\draw[->] (q1.south) -- (q2.north);
\draw[->] (q2.south) -- (q3.north);
\draw[->] (q3.south) -- (q4.north);

\draw[->] (q1.east) -- (r1.west);

\draw[->] (r1.south) -- (r2.north);
\draw[->] (r2.south) -- (r3.north);
\draw[->] (r3.south) -- (r4.north);
\draw[->] (r1.east) -- (5.5, 0);
\draw[->] (-1.5, 0) -- (point.west);
\end{tikzpicture}
\caption{A generic diagram of a deep SSM with $L$ layers. 
%\mred{[THE INDEXING OF THE LATENT PROCESSES NEEDS TO BE CHANGED. THE CLOSEST TO $y$ SHOULD BE $x_L$. THE ROOT PROCESS IS $x_0$.][Done]}
}
\end{figure}
One of the advantages of deep structures is to use one or a few simple nonlinear activation functions to improve the approximation of unknown highly nonlinear target functions. In the field of signal processing, the advantage of deep structures of state-space models is that with more hidden layers we can improve the modeling capacity of the model.
Typically, the state processes of the hidden layers will be of different dimensions, and in some settings, the deep models can be justified using arguments that reflect our understanding of the phenomena we model. A generic diagram of a deep SSM with $L$ layers is shown in Fig. 2.  

Borrowing from concepts of deep learning, we introduce a Gaussian process-based deep state-space model (GP-DSSM). This model uses one simple kernel that is combined with a deep structure to approximate the unknown target kernel. Formally, we express a GP-DSSM with $L$ hidden layers as follows:
\begin{align}
\label{eq:root}
\xx_{1,t} &= \HH_1^\top\bfphi^x_{1,t-1} + \uu_{1,t},
\\
\xx_{l,t} &= \HH_{l}^\top\bfphi_{l{-1},t}^x + \uu_{l,t},\;\;l=2\ldots, L,\\
\yy_t &= \bfTheta^\top\bfphi_{t}^y + \vv_{t},
\end{align}
where $\xx_{l,t} \in \mathbb{R}^{d_x^l}$, $l= 1, 2, \ldots, L$ are latent processes, $\yy_t \in \mathbb{R}^{d_y}$ is a vector of observations,  $\bfphi^x_{l,t}=\bfphi(\xx_{l,t})$ are the feature functions embedded with different $\bfOmega_l$ for every layer, $\bfphi_{t}^y$ has the same meaning as before, ${\bf H}_l$ and $\bfTheta$ are parameter variables, and $\uu_{l,t}$ and $\vv_{t}$ are perturbations.
We refer to the deepest latent process (defined by \eqref{eq:root}) as the root process of the model. 

Here we assume that the dimensions of the latent processes are predefined. The objective of inference is to estimate all the latent processes $\xx_{l,t}$, $l=1, \ldots, L$ and all the parameters of the model $\HH$ and $\bfTheta_l$, $l=1, 2, \ldots, L,$. 

The inference method and procedures are very similar to the method we described for the ordinary GP-SSM. 

\subsection{Propagation of the particles in all layers}
At time $t$, first we propagate the particles from $\xx_{1,t-1}^{(m)}$ to $\xx_{1,t}^{(m)}$ and then the particles of the remaining latent processes $\{\xx_{l,t}^{(m)}\}_{l=2}^{L}$. In propagating these particles, we apply analogous Student's $t$-distributions as in \eqref{eq:linear_pred}, i.e.,
\begin{align}
&p(x_{l,t}^{(m),[d]}|\X_{t-1}^{(m)}, \Y_{t-1}) \notag\\&
\propto \left( 1+\frac{1}{\psi_{l,t}^{(m),[d]}}\left(x_{l,t}^{(m),[d]} - \alpha_{l,t}^{(m),[d]}\right)^2 \right)^{-\frac{\nu_{l,t-1}^x+1}{2}},
\label{eq:transit_unknowns-1nn}
\end{align}
where $\X_{t-1}^{(m)}$ represents the latent states in all the layers up to time $t-1$,  and where the parameters $\nu_{l,t}, \psi_{l,t}$, and $\alpha_{l,t}$ are defined similarly as in \eqref{eq:parameters_unknown-nu} -- \eqref{eq:upPsi}.

\subsection{Updating of the joint posteriors of $(\bfeta_l^{(m),[d]}, \sigma_{l,v}^{{(m),[d]}^2})$ and $(\ttheta^{(m),[d]}, \sigma_v^{{(m),[d]}^2})$}
These updates follow the schemes described by \eqref{eq:aparamx}--\eqref{eq:eta1} and \eqref{eq:aparam_2}--\eqref{eq:upUpsilon_2}, respectively. We note that these updates can be performed in parallel once all the particles in all the layers have been propagated.

\subsection{Weight computation of particles and estimation of the latent processes}

We assign  weights to the particles $\xx_{L,t}^{(m)}$ according to the likelihoods of the particles, that is, we use
\beq
\label{eq:weight_pf-2}
\widetilde{w}_{L,t}^{(m)} =  p(\yy_t|\xx_{L,t}^{(m)}, \X_{t-1}^{(m)},\Y_{t-1}),
\eeq
where $p(\yy_t|\xx_{L,t}^{(m)}, \X_{t-1}^{(m)}, \Y_{t-1})$ is the likelihood of $\xx_{L,t}^{(m)}$ given $\yy_t$, $\X_{t-1}^{(m)},$ $\Y_{t-1}$, and  $\widetilde{w}_{L,t}^{(m)}$ is the non-normalized weight of ${\bf x}_{L,t}^{(m)}$. The computation of this weight is carried out via a Student's $t$-distribution
of the form as in \eqref{eq:parameters35} and whose parameters are from expressions analogous to  \eqref{eq:parameters_nu}--\eqref{eq:upsi}. Upon the computation of the weights, we normalize them as per \eqref{eq:weightnorm}. Clearly, these weights { directly} depend on $\xx_{L,t}^{(m)}$ only and not on the particles from the previous layers. Finally, the minimum mean square estimate (MMSE) of $\xx_{L,t}$ is computed by \eqref{eq:mmse_pf-1} and the approximation of the posterior $p(\xx_{L,t}|\Y_t)$ is given by \eqref{eq:posterior_pf}. 

The estimates of the remaining processes is carried out by first computing the weights of the particles of the corresponding processes. For example, for computing the weights of $\xx_{L-1,t}^{(m)}$, we use 
\beq
\label{eq:weight_pf-2a}
\widetilde{w}_{L-1,t}^{(m)} =  p\left(\widehat{\xx}_{L,t}|\xx_{L-1,t}^{(m)}\right),
\eeq
where $\widehat{\xx}_{L,t}$ is the estimate of ${\xx}_{L,t}$. 
From the particles $\xx_{L-1,t}^{(m)}$ and their corresponding weights, we then compute $\widehat{\xx}_{L-1,t}$.  
We continue in the same vein by estimating one process value at a time until we complete these steps with estimating the value of the root process.   

Before we proceed to process the next observation, we resample the $M$ streams using their respective weights $w_t^{(m)}$. One approach to computing these weights ise based on the following expression:
\beq
\label{eq:weight_pf-3}
\widetilde{w}_{t}^{(m)} =  p\left(\yy_t|\xx_{L,t}^{(m)}, \Y_{t-1}\right)
\prod_{l=1}^{L-1}p\left(\xx_{l+1,t}|\xx_{l,t}^{(m)}\right).
\eeq
For the unknown $\xx_{l+1,t}$ in this equation, %The problem in implementing \eqref{eq:weight_pf-3} is that we do not know the latent processes $\xx_{l,t}$. As a substitute of the unknown $\xx_{l,t}$s, 
we could use their respective MMSE estimates $\widehat{\xx}_{l,t}$. Another approach is based on  approximating the factors $p\left(\xx_{l+1,t}|\xx_{l,t}^{(m)}\right)$ in \eqref{eq:weight_pf-3} with the average likelihood, that is, with 
\begin{align}
\label{eq:mc_approx}
p\left(\xx_{l+1,t}|\xx_{l,t}^{(m)}\right)&\approx \sum_{m{'}=1}^M w_{l+1,t}^{(m{'})}\, p\left(\xx_{l+1,t}^{(m{'})}|\xx_{l,t}^{(m)}\right),
\end{align}
where $w_{l+1,t}^{(m)}$ are the weights associated with the particles $x_{l+1,t}^{(m)}$. 
{By combining equation \eqref{eq:weight_pf-3} and \eqref{eq:mc_approx}, we obtain
\beq
% \label{eq:weight_pf-3-combine}
% \tag{57}
%\notag
\widetilde{w}_{t}^{(m)} \approx p\left(\yy_t|\xx_{L,t}^{(m)}, \Y_{t-1}\right)
\prod_{l=1}^{L-1}\left[\sum_{m{'}=1}^M w_{l+1,t}^{(m{'})}\, p\left(\xx_{l+1,t}^{(m{'})}|\xx_{l,t}^{(m)}\right)\right].
\eeq

\section{Experiments}
We tested the performance of the proposed method with several experiments. In all the experiments, we applied the ensemble method with $S=100$ members. Specifically, the elements of the sets $\{\bfOmega^s\}_{s=1}^{100}$ were randomly sampled from the power spectral density of RBF kernels $\kappa^s(\bflambda)$ with prior length scale vectors $\pmb{l}_\lambda^s$ and prior variances $\sigma_\lambda^2=1$, where the elements of $\pmb{l}_\lambda^s$ were independently sampled from the discrete set $\{10^{-4}, 10^{-3}, \ldots, 10^3, 10^4\}$. {Note that  $\bflambda$ represents the hyperparameters and $\bflambda = \{\pmb{l}_\lambda, \sigma_\lambda^2\}$ includes the length scale vector and the prior variances.} 

\subsection{A test with $d_x=2$ and $d_y=1$}

\begin{figure*}[!htb]
\centering
\begin{minipage}[h]{0.48\textwidth}
\centering
    \includegraphics[width=3.4in]{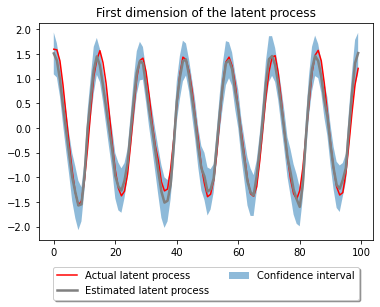}
    \caption{Point estimates and 95\% confidence region of $\widehat{x}_t^{[1]}$ and its true values $x_t^{[1]}$ for the last 100 samples.}
    \label{fig:r1_x1}
\end{minipage}
\hspace{0.01\linewidth}
\begin{minipage}[h]{0.48\textwidth}
\centering
    \includegraphics[width=3.4in]{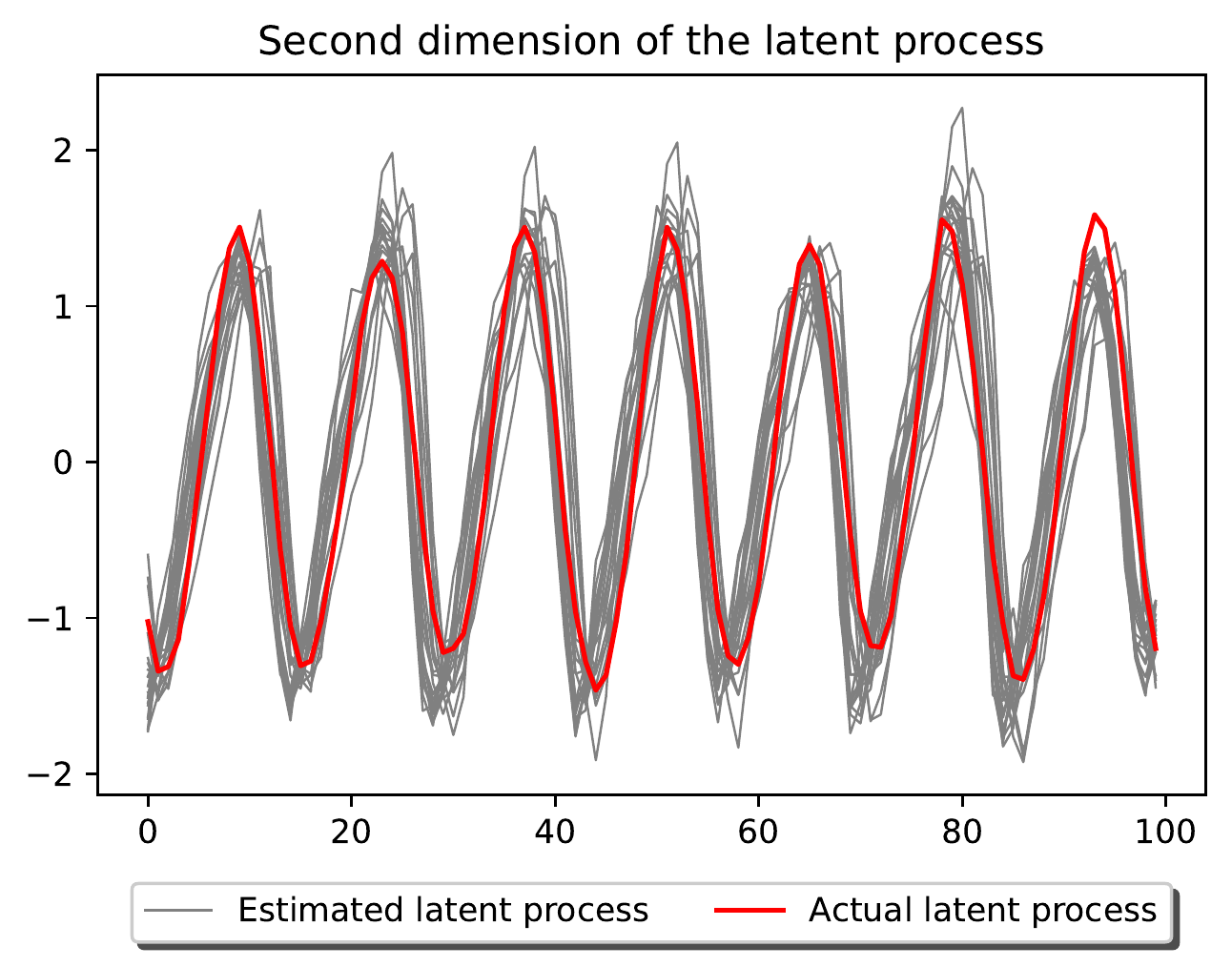}
    %{results_yuhao/result1_x2.pdf}
    \caption{{Estimates of $\widehat{x}_t^{[2]}$ under 20 runs and its true values $x_t^{[2]}$.}}
    \label{fig:r1_x2}
\end{minipage}
\end{figure*}

\begin{figure*}[!htb]
\centering
\begin{minipage}[h]{0.48\textwidth}
\centering
    \includegraphics[width=3.4in]{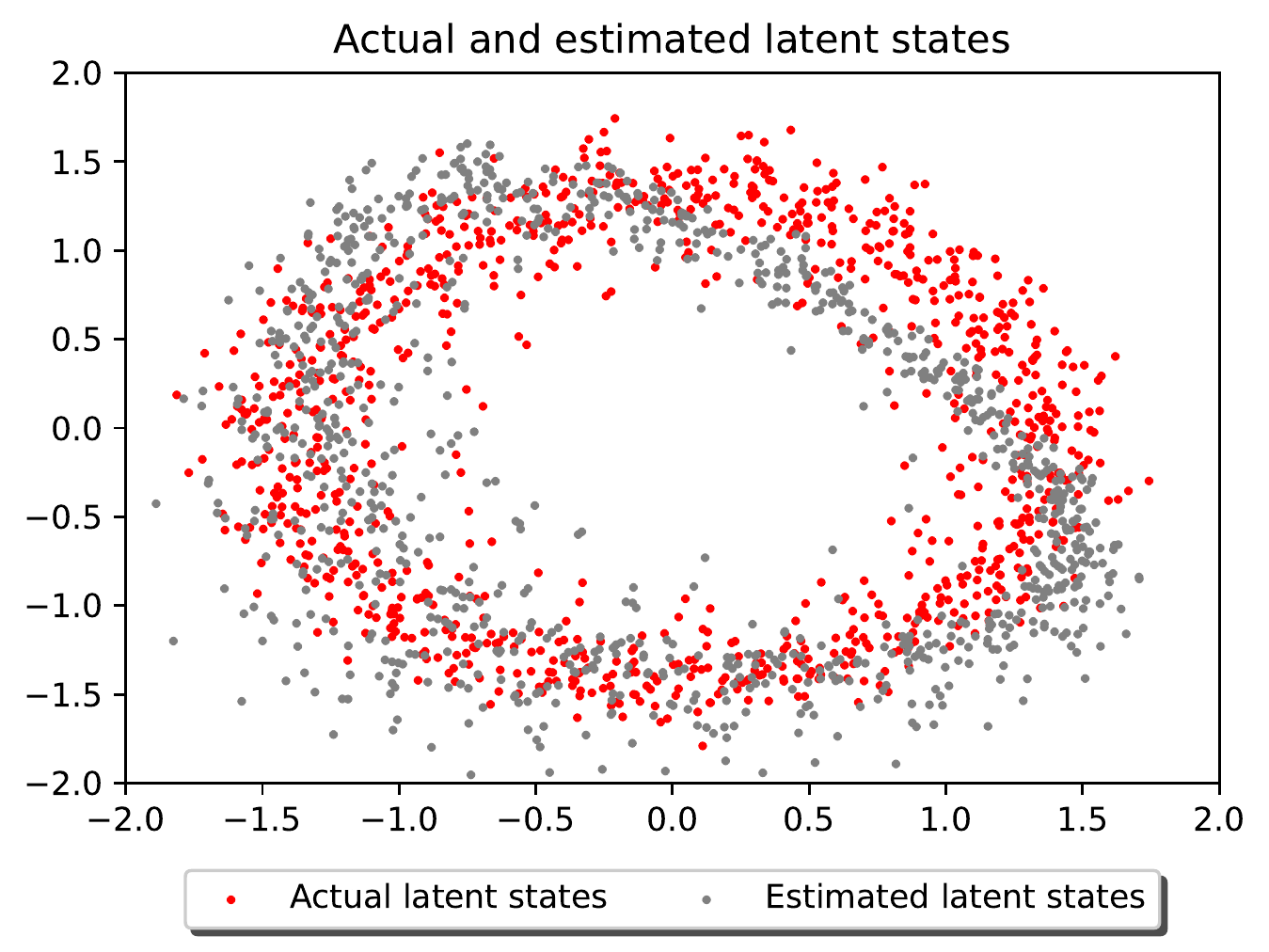}
    \caption{Pairs of estimated ($\widehat{\xx}_t$) and actual ($\xx_t$) latent states before SVD.}
    \label{fig:r1_xhat}
\end{minipage}
\hspace{0.01\linewidth}
\begin{minipage}[h]{0.48\textwidth}
\centering
    \includegraphics[width=3.4in]{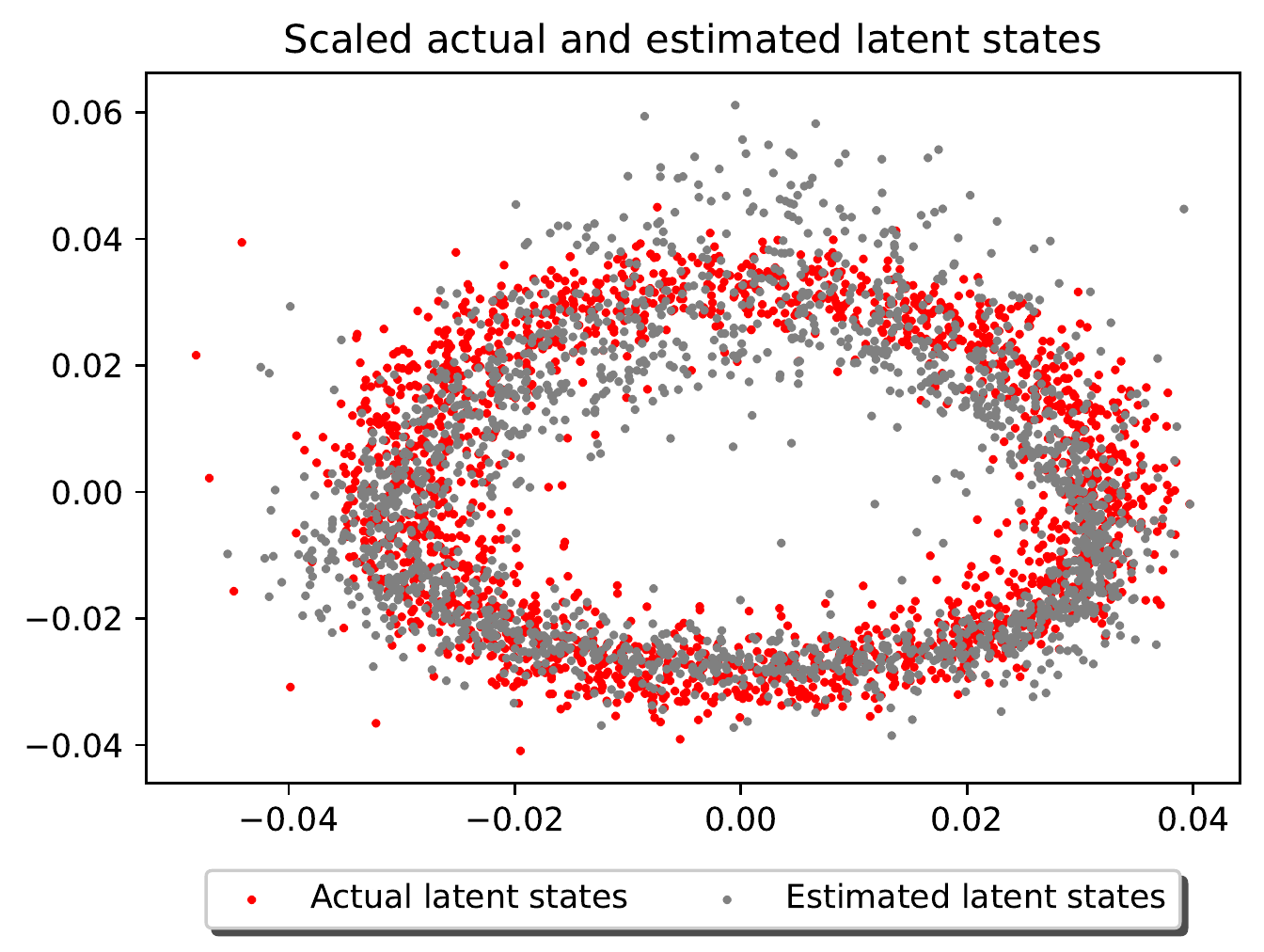}
    \caption{Pairs of standardized $\widehat{\xx}_t$ and $\xx_t$ after SVD.}
    \label{fig:r1_xhat_scaled}
\end{minipage}
\end{figure*}

\begin{figure*}[!htb]
\centering
\begin{minipage}[h]{0.44\textwidth}
\centering
    \includegraphics[width=\textwidth]{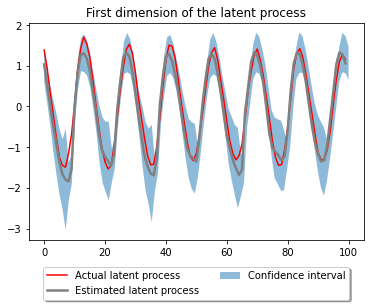}
    \caption{Point estimates of $\widehat{x}_t^{[1]}$ with 95\% confidence region and the true $x_t^{[1]}$ from $T_0$ to $T_0 + 100$.}
    \label{fig:100}
\end{minipage}
\hspace{0.01\linewidth}
\begin{minipage}[h]{0.44\textwidth}
\centering
    \includegraphics[width=\textwidth]{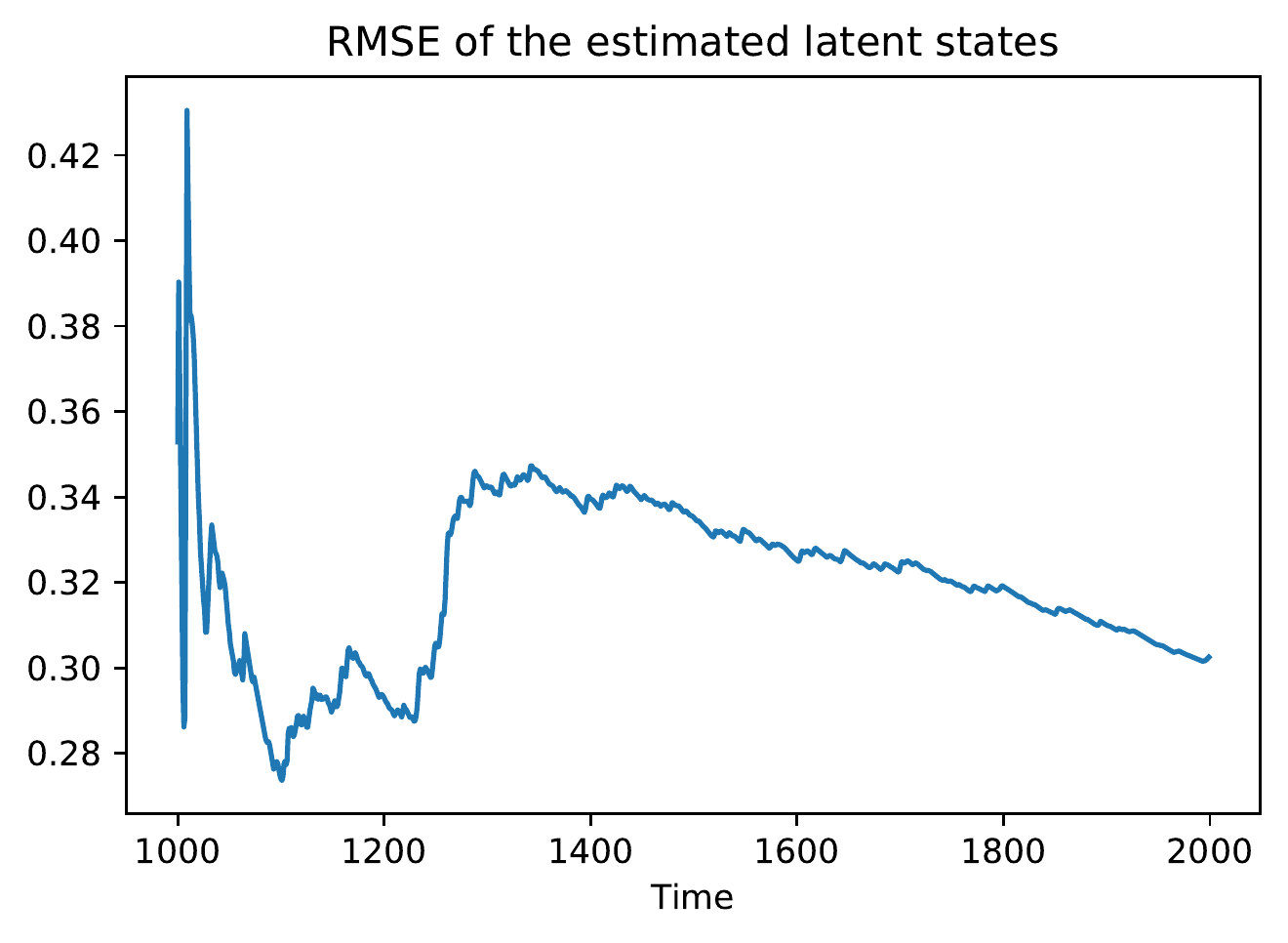}
    \caption{RMSEs of the estimated latent states from $T_0$ to the end.}
    \label{fig:r1_rmse}
\end{minipage}
\end{figure*}

\label{sec:test1}
In the first experiment, we tested the inference of a GP-SSM when $d_x > d_y$. More specifically, we generated data from an SSM with $d_x=2$ and $d_y=1$ according to the following model:
\begin{equation}
\nonumber
\begin{array}{lll}
\smaller
{\rm latent \; layer:}&x_t^{[1]} = 0.9  x_{t-1}^{[1]} + 0.5 \sin(x_{t-1}^{[2]}) + u_t^{[1]},\\
&x_t^{[2]} = 0.5  \cos(x_{t-1}^{[1]}) + 0.9 x_{t-1}^{[2]} + u_t^{[2]},\\\\
{\rm observations:}&y_t = 0.3 \sin(x_t^{[1]}) -0.3 x_t^{[1]} + 0.2 x_t^{[2]} \\
& + 0.25x_t^{[1]} x_t^{[2]} + (0.05x_t^{[1]})^2 + 0.01(x_t^{[1]})^3\\
& - 0.25x_t^{[1]}/(1+(x_t^{[2]})^2) + v_{t}.
\end{array}
% \label{eq:dssm}
\end{equation}
%where we used some commonly used non-linear functions. 
The generated data set contained 2,000 samples with $\sigma_u^2=\sigma_v^2=0.001$, and for initializing the estimation, we used  $T_0=1,000$ samples. For drawing random vectors needed in the construction of the random features $\bfphi(\xx)$ in \eqref{rf_vector}, we used $J_x=J_y=50$.
For comparison purposes, all the signals were normalized from $T_0=1,000$ to $T=2,000$. We emphasize again that the functions in the state and observation equations are unknown. 
{Figure \ref{fig:r1_x1} shows the last $100$ samples of the actual latent process $x_t^{[1]}$ in red line, its estimate $\widehat{x}_t^{[1]}$ in grey line, and the 95$\%$ confidence interval depicted by the blue region. Figure \ref{fig:r1_x2} exhibits the estimates of the latent states $\widehat{x}_t^{[2]}$ under 20 runs with different seeds.}
The results suggest that with our approach we can provide accurate estimates of the latent states and thus can capture their dynamics. Further, we can quantify the uncertainties of the estimates.
Recall from Section \ref{sec:ens_states} that we use the SVD  to standardize both the actual and estimated states.
Figure \ref{fig:r1_xhat} illustrates the true pairs $(x_t^{[1]}, x_t^{[2]})$ and estimated pairs $(\widehat{x}_t^{[1]}, \widehat{x}_t^{[2]})$ before applying SVD.  The scaled actual and estimated states after SVD are shown in Fig. \ref{fig:r1_xhat_scaled}. 

{To assess the performance of our method further, 
in Fig. \ref{fig:100} we present the results of the first 100 samples, demonstrating the consistent performance of our method. It is important to note that the samples from $T_0$ to the end have been standardized simultaneously, ensuring a consistent standardization approach across the entire test set. Thus, the first and last 100 samples have not been separately standardized. In addition, in Fig. \ref{fig:r1_rmse} we show the root mean square errors (RMSEs) of the estimated latent states.}

{Here we provide motivation and insights for using Student's t-distributions rather than Gaussian ones. In our previous work, \cite{liu2022inference}, we considered the variances of the Gaussians $\bfsigma^2$ as vectors that are optimized by gradient descent algorithms. However, bad initial values of variances would incur huge bias because of the risk of not converging. Therefore, we used  Student's t-distributions to account for the uncertainty of the variances $\bfsigma^2$. Further, the Student's t-distribution allows for a closed-form formulation of the variance updates. To make a comparison between the models with Student's t  and Gaussian distributions, we conducted the following experiment. We assumed that the prior information provides initial values of Gaussian variances with 0.1, while the actual variances were $\sigma_u^2=\sigma_v^2=0.001$. The model with Gaussians had a much worse performance in accuracy and had increased computing time compared to the model with Student's t distributions. The results are shown in Figs.  \ref{fig:t_rmse} and \ref{fig:t_mnll}. The red lines show the RMSEs and MNLLs under the model with Gaussians, whereas the grey lines represent our proposed model with Student's t distributions. We reiterate that the model with Gaussians requires more time to run for the same number of samples.}

\begin{figure*}[!htb]
\centering
\begin{minipage}[h]{0.48\textwidth}
\centering
    \includegraphics[width=\textwidth]{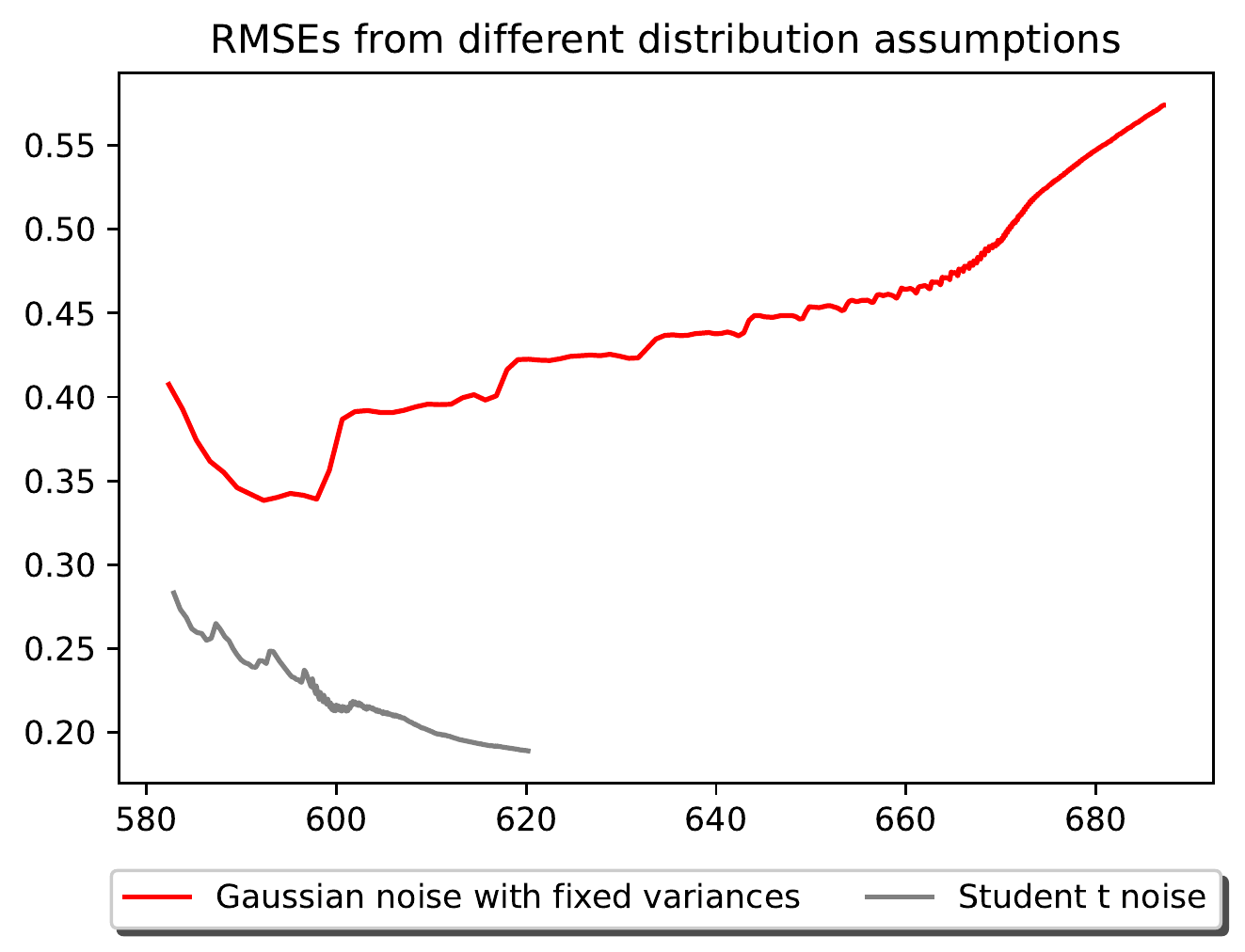}
    \caption{{RMSEs after $T_0$. The x-axis represents time (seconds).}}
    \label{fig:t_rmse}
\end{minipage}
\hspace{0.01\linewidth}
\begin{minipage}[h]{0.48\textwidth}
\centering
    \includegraphics[width=\textwidth]{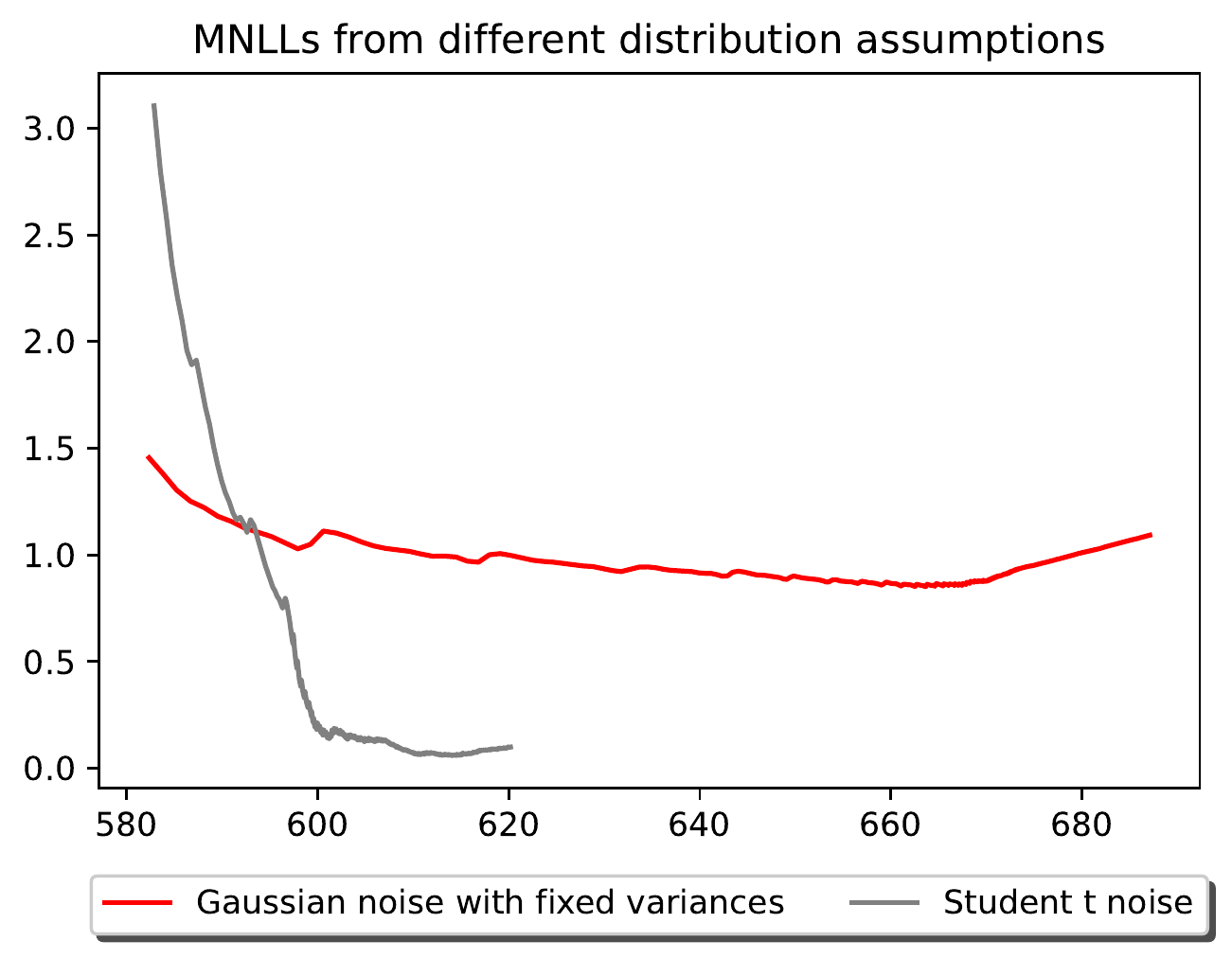}
    \caption{{MNLLs after $T_0$. The x-axis represents time (seconds).}}
    \label{fig:t_mnll}
\end{minipage}
\end{figure*}

\subsection{A test with $d_x=5$ and $d_y=100$}
In the next experiment, we tested the GP-SSM when $d_x < d_y$. We wanted to demonstrate the ability of our model to learn lower dimensional processes from high dimensional observation signals. Our generative model had %\st{$d_x=1$ and $d_y=10$} 
{$d_x=5$ and $d_y=100$} and was of the form
\begin{equation}
\nonumber
\begin{array}{lll}
\smaller
{\rm Latent \; Layer:}&x_t^{{[i]}} = \bfphi_x^\top({\xx_{t-1}})\bfeta^{{[i]}} + u_t^{{[i]}},\\
{\rm Observations:}&y_t^{[j]} = \bfphi_y^\top({\xx_{t}})\bftheta^{[j]} + v_{t}^{[j]},
\end{array}
% \label{eq:dssm}
\end{equation}
where {$i=1,\ldots, 5$, $j=1,\ldots, 100$}, 
%\st{$j=1,\ldots, 10$}, 
$\bfphi_x^\top(x) = [\sin(\bfomega_x^\top x) \; \cos(\bfomega_x^\top x)]$ and $\bfphi_y^\top(x) = [\sin(\bfomega_y^\top x) \; \cos(\bfomega_y^\top x)]$. The elements of $\bfomega_x \in \bbR^{50}$ and $\bfomega_y \in \bbR^{50}$ were randomly generated from $-10$ to $10$, and the entries of $\bfeta \in \bbR^{100}$ and $\bftheta \in \bbR^{100}$ were also randomly drawn from $-0.01$ to $0.01$. The hyperparameters were set to be the same as in the above section. Figure \ref{fig:high_dim} shows 
%\st{$x_t$ and $\widehat{x}_t$}
{$\xx_t$ and $\widehat{\xx}_t$} of the last $100$ samples. 
%\st{The results indicate that even when the signal shrinks and then enlarges suddenly, our model can adjust quickly and precisely.} 
{The results indicate that even for the signals with high frequency and dimensions, our model can adjust quickly.}

\begin{figure*}[!htb]
  \begin{center}
  \includegraphics[width=\textwidth]{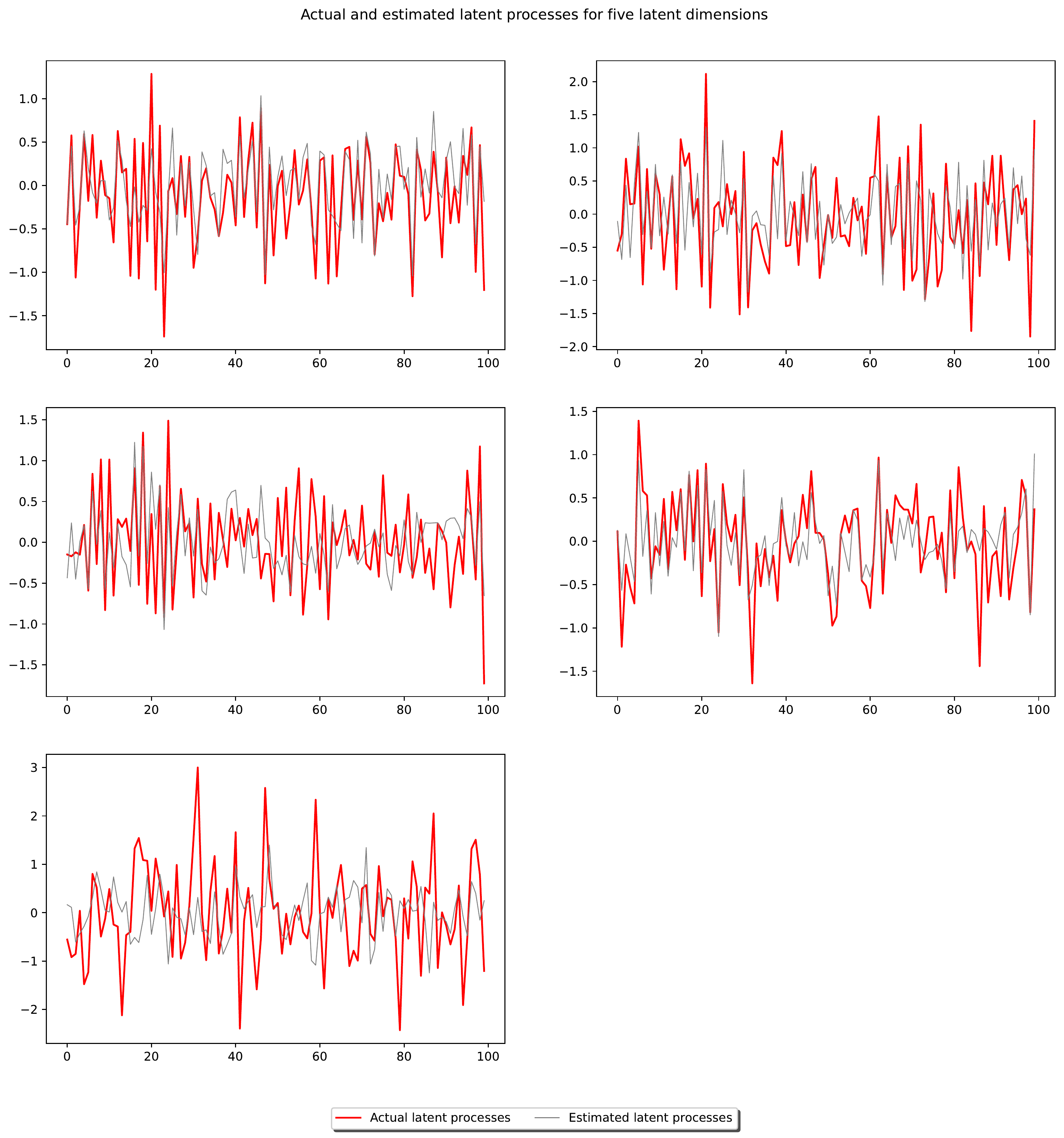}
  \caption{{Estimated latent processes $\widehat{\xx}_t$ and actual latent processes $\xx_t$ for all five dimensions.}}
  \label{fig:high_dim}
  \end{center}
\end{figure*}

\subsection{The need for a deep model}
Do we need GP-DSSMs? This experiment shows that the answer is positive, especially when the selected kernel for the GP may not have enough capacity to learn.
We validated this by an experiment where $d_x=10$ and $d_y=1$.
We generated the observation process as a GP whose kernel was a superposition of a dot-product and a Mat\'ern kernel.
%\st{and a white kernel}. 
The dot-product kernel was a non-stationary kernel with a hyperparameter $\sigma_{dp}^2=20$, 
and the Mat\'ern kernel was set with hyperparameters $\nu=1.5$ and length scales $\pmb{l}=[10^{-5},10^{-4},\ldots,10^3,10^4]$. 
The outputs were normalized before being used by our model. In mathematical terms, the {transition and observation processes were obtained by }
\begin{align}
x_t^{[i]} &= 0.9x_{t-1}^{[i]} + g_i(x_t^{[i+1]})/2 + u_t^{[i]},\\
y_t &= f(x_t) + v_t,
\end{align}
where $i=1,\ldots,10$, and $x_t^{[11]}$ actually denotes $x_t^{[1]}$ to simplify the notation. The function 
$f$ is the GP with a dot-product kernel 
adding a Mat\'ern kernel, and $g_i$ is a sine function when $i$ is odd while a cosine when $i$ is even. 
The noises $u_t^{[i]}$ and $v_t$ had the same variances $\sigma_u^2=\sigma_v^2=0.1$.
The remaining parameters were $J_x=J_y=100$ and $M=10^4$.

We implemented four models, from one-hidden layer to four-hidden layers. Figure \ref{fig:r4_rmse} shows the RMSEs of the four models. The model with two-hidden layers achieves the minimum RMSEs. Note that the behavior of the RMSEs relative to the number of hidden layers is similar to a concave function, consistent with the conclusion from \cite{cutajar2017random}, i.e., that the RMSEs decrease and then increase with the number of hidden layers increasing. From the conclusion in \cite{dunlop2018deep}, we might expect that deeper models would be better when we increase the number of parameters such as $J$ and $M$.

\begin{figure}[htb]
  \begin{center}
  \includegraphics[width=3.5in]{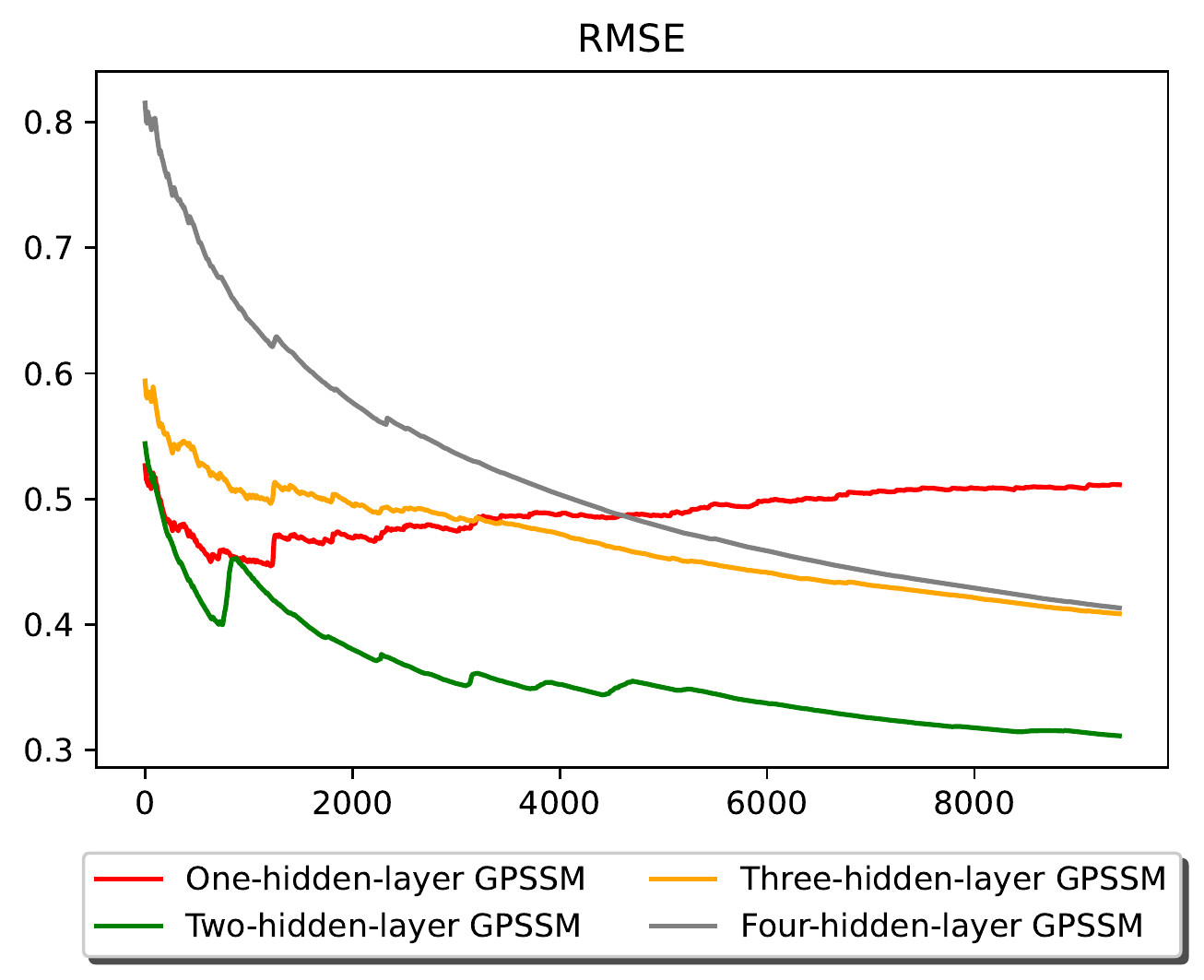}\\
  \caption{{RMSEs obtained from a single-hidden-layer GP-SSM to a four-hidden-layer GP-DSSM.}}\label{fig:r4_rmse}
  \end{center}
\end{figure}

\subsection{Testing the performance of GP-DSSM}

\begin{figure*}[!htb]
\centering
\includegraphics[width=\textwidth]{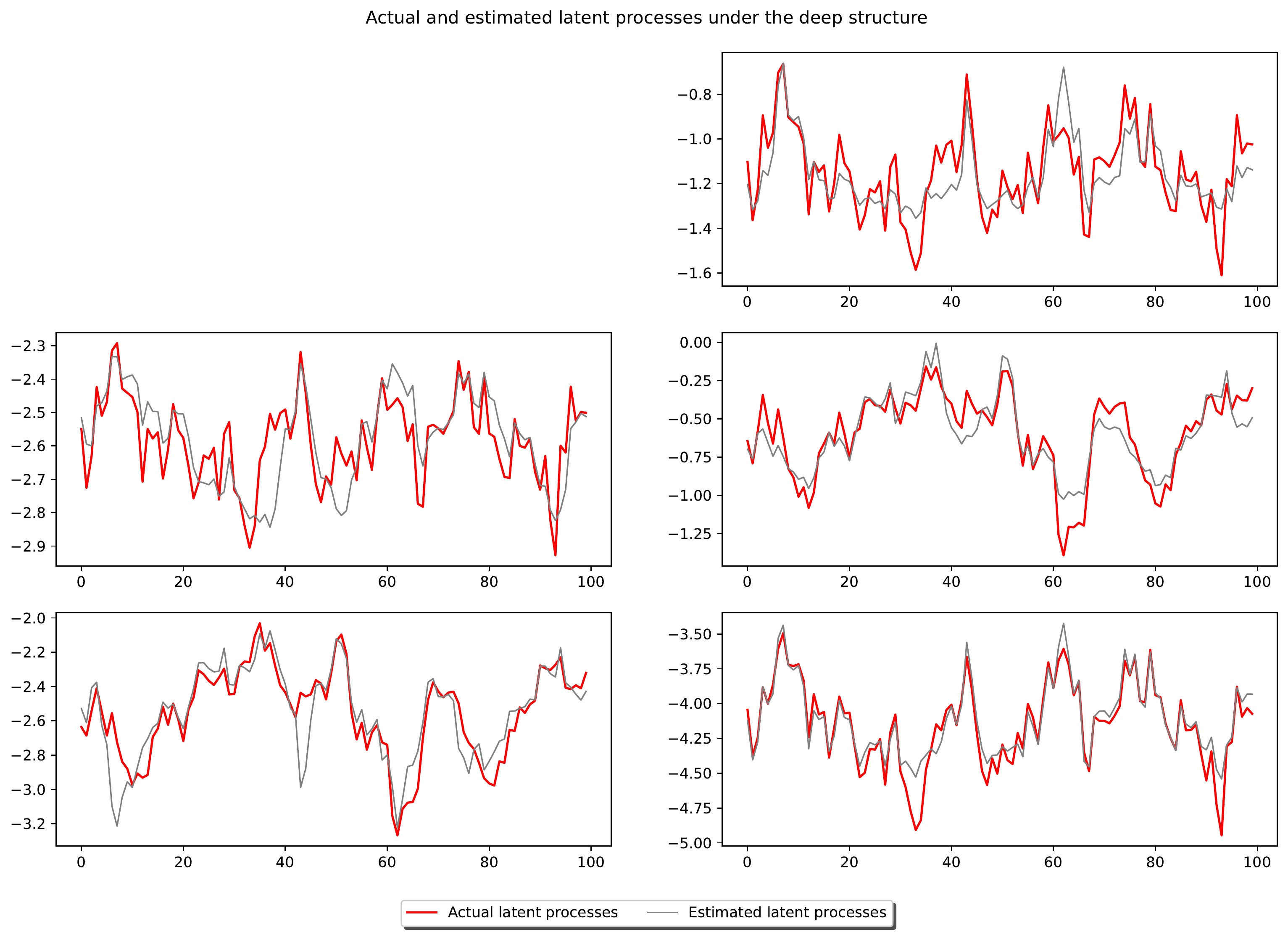}
\caption{On the left are the true values and estimates of $\xx_{1,t}$, and on the right, the true values and estimates of $\xx_{2,t}$.}
%are the true values and the estimates of the proposed method.}
\label{fig:DSSresults}
\end{figure*}

We generated data from a DSS model with two hidden layers,  with $d_{x_1}=2$,  $d_{x_2}=3$, and $d_y=4$. The model is given by the following two layers of latent processes:
\begin{equation}
\begin{array}{lll}
\smaller
{\rm Layer\;1:}&x_{1,t}^{[1]} = 0.9  x_{1,t-1}^{[1]} + 0.5 \sin(x_{1,t-1}^{[1]}) + u_{1,t}^{[1]},\\
&x_{1,t}^{[2]} = 0.5  \sin(x_{1,t-1}^{[1]}) + 0.9 x_{1,t-1}^{[2]}] + u_{1,t}^{[1]},\\\\
{\rm Layer\;2:}&
x_{2,t}^{[1]} = 1.8 \cos(x_{1,t}^{[1]}) - 0.7  \sin(x_{1,t}^{[1]}) + u_{2,t}^{[1]},\\&
x_{2,t}^{[2]} = 0.5  x_{1,t}^{[1]} - 1.3  \sin(x_{1,t}^{[2]}) + u_{2,t}^{[1]},
\\
&x_{2,t}^{[3]} =  2  x_{1,t}^{[1]} - 0.4  x_{1,t}^{[2]} + u_{2,t}^{[2]}, &
\end{array}
\label{eq:dssm_exp}
\end{equation}  
and four observation processes given by 

\begin{equation}
\begin{array}{lll}
\smaller
&y_t^{[1]} = 0.01  (x_{2,t}^{[1]})^2                                  + 1.2  x_{2,t}^{[3]}   + v_{t}^{[1]},\\& 
y_t^{[2]} = 1.2  \sin(x_{2,t}^{[1]})- 0.5  x_{2,t}^{[2]}  + 0.7  x_{2,t}^{[3]} + v_{t}^{[2]} ,\\
&y_t^{[3]} = x_{2,t}^{[1]}  x_{2,t}^{[2]}   
+  v^{[3]}_{t},  \\
&y^{[4]}_{t} = 5x_{2,t}^{[2]} / (1 + x_{2,t}^{{[2]}^2}) + v^{[4]}_{t}.
\end{array}
\label{eq:dssm_exp2}
\end{equation}  

This model is identical to the one from Sec. V-B of \cite{liu2022inference}, except that we changed the first equation in \cite{liu2022inference}
\beq
x_{1,t}^{[1]} = 0.9  x_{1,t-1}^{[1]} + 0.5 \sin(x_{1,t-1}^{[2]}) + u_{1,t}^{[1]},
\eeq
to
\beq
x_{1,t}^{[1]} = 0.9  x_{1,t-1}^{[1]} + 0.5 \sin(x_{1,t-1}^{[1]}) + u_{1,t}^{[1]}.
\eeq
We made this change to force the latent process to become less smooth, thus making it harder for estimation. The results are shown in Fig. \ref{fig:DSSresults}. Evidently, they show that the proposed method is capable of accurately estimating all the latent processes even though the latent processes are much more jagged. Further, compared with the results in \cite{liu2022inference} which had persistent lags in the estimated processes, the results from the method presented here showed almost no lags. % at the root layer and the intermediate layer.

\subsection{Testing on real-world data sets}
{We also assessed the performance of our model on five real-world data sets \cite{doerr2018probabilistic} and compared it to six state-of-the-art models. The suite of reference methods is composed of (1) two one-step ahead autoregressive GP models: GP-NARX \cite{kocijan2005dynamic} and NIGP \cite{mchutchon2011gaussian}, (2) three multi-step-ahead autoregressive and recurrent GP
models in latent spaces: REVARB with one and two hidden layers \cite{mattos2017deep} and MSGP \cite{doerr2017optimizing}, and (3) two GP-SSMs, based on a full Markovian state:
SS-GP-SSM \cite{svensson2017flexible} and PR-SSM \cite{doerr2018probabilistic}. Specifically, the REVARB is a deep state-space model based on RNNs. Note that these benchmark methods are learned in an offline mode or in a batch mode, which means that they are trained on sets multiple times and then applied to test sets.

\begin{table*}[!htbp]
    \caption{RMSEs and standard deviations}
    \label{tab:real_data}
    \centering
    \begin{tabular}{r c c c c c c c c c p{0.3\textwidth}}
    \toprule
     & \multicolumn{2}{c}{ONE-STEP-AHEAD,} &\multicolumn{3}{c}{MULTI-STEP-AHEAD, LATENT} &\multicolumn{3}{c}{MARKOVIAN STATE-SPACE}\\
     & \multicolumn{2}{c}{AUTOREGRESSIVE} & \multicolumn{3}{c}{SPACE AUTOREGRESSIVE} & 
     \multicolumn{3}{c}{MODELS}\\
    \cmidrule(r{1em}){2-3}\cmidrule(r{1em}){4-6}\cmidrule(r{1em}){7-9}
    TASK & GP-NARX & NIGP & REVARB 1 & REVARB 2 & MSGP & SS-GP-SSM & PR-SSM & RF-SSM \\
    \midrule
    ACTUATOR & 0.627  & 0.599  & 0.438  & 0.613  & 0.771  & 0.696  & 0.502  & \textbf{0.295 } \\
    & (0.005) & (0) & (0.049) & (0.190) & (0.098) & (0.034) & (0.031) & \textbf{(0.037)} \\
    BALLBEAM & 0.284  & 0.087 & 0.139  & 0.209  & 0.124  & 411.6 & \textbf{0.073} & 0.107 \\
    & (0.222) & (0) & (0.007) & (0.012) & (0.034) & (273.0) & \textbf{(0.007)} & (0.010) \\
    DRIVE & 0.701 & \textbf{0.373 } & 0.828  & 0.868 & 0.451  & 0.718  & 0.492  &  0.417 \\
    & (0.015) & \textbf{(0)} & (0.025) & (0.113) &  (0.021) & (0.009) & (0.038) &  (0.030) \\
    FURNACE & 1.201 & 1.205  & 1.195 & 1.188  & 1.277  & 1.318  & 1.249  & \textbf{0.410 } \\
    & (0.000) & (0) &  (0.002) &  (0.001) &  (0.127) & (0.027) &  (0.029) & \textbf{ (0.032)} \\
    DRYER & 0.310  & 0.268  & 0.851 & 0.355  & \textbf{0.146} & \textbf{0.152 } & \textbf{0.140 } & 0.273 \\
    & (0.044) &  (0) &  (0.011) &  (0.027) & \textbf{ (0.004)} & \textbf{(0.006)} & \textbf{ (0.018)} & (0.021) \\
    \bottomrule
    \end{tabular}
\end{table*}

The settings of all the methods were the same, where the number of inducing points $P$ or the number of random features $J$ was 20. The latent dimension $d_x$ was set to four. All the observations were normalized. The results of the best performer in terms of the Welch t-test are presented in bold numbers. Further information about our model including run time is provided in Table \ref{tab:data_information}. The training times were collected from a server with 12G RAM. The training time of our ensemble model depends on the number of candidate models, which was $S=100$ in our experiments, and these models were trained separately. If they are deployed in a distributed manner, the training time in Table \ref{tab:data_information} can be reduced by around 100 times. The benchmark methods, however, can only be implemented by multiple iterations and cannot be conducted in a distributed way. The training and test times of the other methods are not provided in the corresponding papers and are affected by the used number of iterations in the computations.

As benchmarks, we have chosen to use batch methods due to the lack of other sequential methods that operate under the assumption of unknown transition and observation functions, as described in our paper. In order to ensure a fair comparison, we utilized a test set with a small number of samples, specifically ranging from 100 to 500 samples. Opting for a smaller test set allows for a more gradual and less rapid change in the sequential method. The batch methods undergo multiple training iterations on the training set to ensure convergence of their parameters. By contrast, our method only exploited the training set once and thus, may have not achieved complete convergence by the end of the training process due to the small number of training samples. %\sout{Our proposed model, however, operates in  online or sequential mode and only uses a training set once}}. 
Even so, our method, RF-SSM, achieved the best performance on two data sets (\textit{Actuator} and \textit{Furnace}). It also was the second best performer on the data set \textit{Drive} and the third best performer on the data set \textit{Ballbeam}. All the results are shown in Table \ref{tab:real_data}. The numbers in parentheses are the standard deviations of the RMSEs among five runs with different seeds.

\begin{table}[!htb]
    \caption{Data Information}
    \label{tab:data_information}
    \centering
    \begin{tabular}{r c c c c c p{0.25\textwidth}}
    \toprule
    \textit{Task} & $N_{train}$ & $N_{test}$ & $T_{train}(seconds)$ & $T_{test}(seconds)$\\
    ACTUATOR & 512 & 512 & 701.0 (3.847) & 13.8 (0.748) \\
    BALLBEAM & 500 & 500 & 689.6 (15.318) & 20.6 (1.744)\\ 
    DRIVE & 250 & 250 & 307.6 (3.2) & 4.2 (0.400) \\ 
    FURNACE & 148 & 148 & 160.4 (0.490) & 3.8 (0.400)\\ 
    DRYER & 500 & 500 & 686.0 (6.841) & 8.0 (0.00)\\ 
    \bottomrule
    \end{tabular}
\end{table}

\section{Conclusions}
In this paper, we addressed the problem of sequential estimation of state-space models and deep state-space models using Gaussian process-based state-space modeling and Gaussian process-based deep state-space modeling. {An important advantage of the considered methodology is that it relaxes the assumption of knowing the functions in the observation and state equations.} We implemented the Gaussian processes by using random feature-based Gaussian processes. The inference method is based on the combination of particle filtering and Bayesian linear regression. We also proposed an ensemble of filters for tracking the latent processes. With several experiments, we demonstrated the performance of the proposed method in different settings including synthetic examples and five real data sets. Further, we compared the performance of our method with 6 other state-of-the-art methods. A limitation of our approach is that for too deep models, the method requires many more particles and a larger number of features. In future work, we will address the use of the variational Bayes approach to acquire a better set of random feature candidates so that we can reduce the number of features and particles. We will also consider applying more efficient sampling approaches.

\bibliographystyle{IEEEtran}
\bibliography{additional,petar-refs,medium,GP,Bibliography}

\end{document}